\documentclass[preprint,12pt]{elsarticle}

\usepackage{graphicx}
\usepackage{amssymb}
\usepackage{mathrsfs}
\usepackage{amsmath}
\usepackage{amsfonts}

\usepackage{lineno}

\usepackage{graphicx}
\usepackage[utf8]{inputenc}
\usepackage{lmodern}
\usepackage[T1]{fontenc}
\usepackage{lscape}
\usepackage{amsmath}	
\usepackage{amssymb}
\usepackage{cases}
\usepackage{graphicx}
\usepackage{caption}
\usepackage{comment}

\usepackage{makecell}
\usepackage{multirow}
\usepackage{amsfonts}
\usepackage{float} 
\usepackage{diagbox}
\usepackage{fullpage}
\usepackage{url}
\usepackage{rotating}
\usepackage{eurosym}
\usepackage{wrapfig}
\usepackage[final]{pdfpages} 
\usepackage{epstopdf} 
\usepackage[a4paper]{geometry}
\usepackage[nottoc]{tocbibind}
\usepackage{placeins}
\usepackage{float}
\usepackage{listings}
\usepackage{color, colortbl}
\usepackage{hyperref}
\usepackage{xcolor}
\usepackage{graphicx, caption, subcaption}
\usepackage[nohyperlinks]{acronym}
\usepackage{sidecap}
\usepackage{algorithm,algpseudocode}
\usepackage{multirow}
\usepackage{booktabs}
\hypersetup{
colorlinks,
citecolor=black,
filecolor=black,
linkcolor=black,
urlcolor=black
} 

\journal{Computer Methods in Applied Mechanics and Engineering}
\usepackage{svg}

\begin{document}
\begin{frontmatter}


\title{Dynamical system prediction from sparse observations using deep neural networks with Voronoi tessellation and physics constraint}

\author{Hanyang Wang$^{1}$, Hao Zhou$^{2}$, Sibo Cheng$^{3,*}$}

\address{   \small $^{1}$ School of Mathematical Sciences, Faculty of Science, University of Nottingham, UK\\
      \small $^{2}$ School of Mechanical, Medical and Process Engineering, Faculty of Engineering, Queensland University of Technology, Australia\\
      \small  $^{3}$ CEREA, \'{E}cole des Ponts and EDF R\&D, \^Ile-de-France, France\\
      \small  $^{*}$ Corresponding author:  sibo.cheng@enpc.fr\\
}



\begin{abstract}
Despite the success of various methods in addressing the issue of spatial reconstruction of dynamical systems with sparse observations, spatio-temporal prediction for sparse fields remains a challenge. Existing Kriging-based frameworks for spatio-temporal sparse field prediction fail to meet the accuracy and inference time required for nonlinear dynamic prediction problems. In this paper, we introduce the Dynamical System Prediction from Sparse Observations using Voronoi Tessellation (DSOVT) framework, an innovative methodology based on Voronoi tessellation which combines convolutional encoder-decoder (CED) and long short-term memory (LSTM) and utilizing Convolutional Long Short-Term Memory (ConvLSTM). By integrating Voronoi tessellations with spatio-temporal deep learning models, DSOVT is adept at predicting dynamical systems with unstructured, sparse, and time-varying observations. CED-LSTM maps Voronoi tessellations into a low-dimensional representation for time series prediction, while ConvLSTM directly uses these tessellations in an end-to-end predictive model. Furthermore, we incorporate physics constraints during the training process for dynamical systems with explicit formulas. Compared to purely data-driven models, our physics-based approach enables the model to learn physical laws within explicitly formulated dynamics, thereby enhancing the robustness and accuracy of rolling forecasts. Numerical experiments on real sea surface data and shallow water systems clearly demonstrate our framework's accuracy and computational efficiency with sparse and time-varying observations. 
\end{abstract}

\end{frontmatter}

\newpage
\section*{Main Notations}
\begin{table*}[ht!]
    \centering
    \begin{tabular}{ p{3.5cm} p{13.5cm} }
        $T$ & The total time steps of fields. \\
        $N_{\text{epoch}}, N_{\text{init}}$ & Number of epochs, and initial MSE-focused epochs for ConvLSTM, respectively.\\
        $\eta$ & Learning rate for the DSOVT training process.\\
        $N_x, N_y, N_c$ & Space dimensions: spatial ($N_x$, $N_y$), and channel number. \\
        $n_c$ & Index of a channel in dynamical systems. \\
        $k \in \{1, \ldots, K\}$ & Index for each of $K$ sensors. \\
        $(i_{t,k}, j_{t,k})$ & $k^{th}$ sensor position at time $t$. \\
        $R_{t,k}$ & Voronoi cell associated with the $k^{th}$ sensor at time $t$. \\
        $o_{t,k}, \tilde{o}_{t, R_{t,k}}$ & Observed value at the position of the $k^{th}$ sensor and the observed value assigned to the sensors in $R_{t,k}$, respectively.\\
        $\mathbf{x}_t, \mathbf{x}^r_t, \hat{\mathbf{x}}^{CLSTM}_{t}$ & State field in the entire space, tessellated observation field, and predicted field by ConvLSTM at time $t$. \\
        $\tilde{\mathbf{x}}_t$ & Full observation field at time t.\\
        $\mathscr{F}_e, \mathscr{F}_d$ & Encoding and decoding functions of the CED, respectively.\\
        $\theta_{e}, \theta_{d}$ & Parameters of the encoder and decoder, respectively.\\
        $\theta_{LSTM}, \theta_{CLSTM}$ & Parameters of the LSTM and ConvLSTM, respectively. \\
        $\mathbf{h}_t, \hat{\mathbf{h}}_t$ & Latent space representation of Voronoi tessellation via CED and its prediction from an LSTM model at time $t$. \\
        $Z$ & Dimension of latent representation of the CED model\\
        $u, v, h, r$ & Fluid height field ($h$), horizontal ($u$), vertical ($v$), and radius components ($r$) of the shallow water field, respectively.\\
        $\frac{\partial h}{\partial x}, \frac{\partial h}{\partial y}$ & Spatial derivatives of the height field $h$ in the $x$ and $y$ directions. \\
        $\frac{\partial u}{\partial x}, \frac{\partial v}{\partial y}$ & Spatial derivatives of the velocity components $u$ and $v$ in the $x$ and $y$ directions. \\
        $S_{\text{in}}, S_{\text{out}}$ & Input and output steps for our temporal models.\\
        $\mathbf{e}_{\text{in}}, \mathbf{e}_{\text{out}}$ & Input and output energy, respectively.\\
        $\hat{\mathbf{x}}^{CED}_{t}$, $\hat{\mathbf{x}}^{LSTM}_{t}$& Reconstructed field via CED and predicted field by LSTM, respectively. \\
        $\mathcal{E}$ & Energy calculation function. \\
        $\mathcal{L}_{CED},\mathcal{L}_{total}$ & Loss of CED and composite loss of temporal models. \\
        $\mathcal{L}_{\text{energy}}$ & Loss functions for energy conservation. \\
        $\lambda_{energy}$ & Weighting coefficients for energy conservation loss. \\
    \end{tabular}
\end{table*}
\clearpage

\section{Introduction}
The spatio-temporal prediction of nonlinear dynamical systems is essential for real-time decision-making in various fields including engineering and science, with applications in traffic management systems~\cite{diao2019dynamic, liu2021multicomponent}, fluid dynamics~\cite{hu2022hope, wu2023deep}, agricultural practices~\cite{switzman2015modeling}, and atmospheric sciences~\cite{nalli2011multiyear, muduli2016subspace}. Predictions in dynamical systems often encounter challenges due to limited data availability, sparse and unstructured observations, and dynamic sensor placements~\cite{cheng2024efficient}, resulting in gaps in spatial and temporal data coverage that compromise the accuracy and applicability of predictive models~\cite{fattahi2019learning, brunton2016discovering}.

Traditional statistical algorithms such as \ac{VAR} models effectively process temporally rich but spatially sparse data~\cite{Luna2005Predictive}. Sparse regression and data fusion techniques improve prediction accuracy by integrating spatio-temporal factors~\cite{zheng2020short, zhao2018integrated}. However, these methods are sensitive to parameter selection and face challenges due to high computational demands~\cite{Smirnov2005Computation, suesse2018estimation, wu2019dependent, schaeffer2017sparse}. Additionally, the \ac{SINDy} algorithm, which is effective in identifying governing equations from sparse data, faces difficulties when dealing with variable sensor numbers and time-varying positions and is highly sensitive to noise~\cite{brunton2016discovering, zhang2019convergence}.

Recently, there has been increased interest in applying \ac{ML} to dynamical systems~\cite{cheng2023machine}. Researchers are integrating \ac{ROM} with \ac{ML} techniques to address the high computational demands of traditional methods~\cite{wang2021flow, wu2023deep, xiao2019error, xiao2019reduced}. Techniques such as \ac{AE} have proven to be effective in compressing high-dimensional spatial and sequential data for nonlinear dynamics~\cite{badrinarayanan2017segnet, scheinker2021adaptive, gonzalez2018deep}, offering a more efficient solution compared to linear model reduction methods like \ac{POD}. While \ac{POD} is more interpretable, it is less computationally efficient and less suitable for non-linear dynamics~\cite{rahman2019nonintrusive, chatterjee2000introduction}.
To predict temporal dynamics, \ac{LSTM} networks are used~\cite{smagulova2019survey}, with \ac{CAE}-related algorithms reducing computational demands by extracting latent space~\cite{rahman2019nonintrusive,hasegawa2020machine,maulik2021reduced,reddy2019reduced}.
For example, Maulik et al. developed a CAE-LSTM model that effectively manages dynamic evolution in \ac{PDEs}~\cite{maulik2021reduced}, and Masoumi et al. extended this approach to urban airflow dynamics~\cite{masoumi2022improving}. 
The work by Hasegawa et al. (2020)~\cite{hasegawa2020cnn} also demonstrated the effectiveness of the model for unsteady two-dimensional flows around a circular cylinder at different Reynolds numbers.
Furthermore, other temporal latent integrators can be considered as alternatives to LSTM. Maulik et al.~\cite{maulik2021latent} explored the use of Gaussian process emulation for the latent-space time evolution of non-intrusive reduced-order models, providing a probabilistic framework for modeling temporal dynamics. Fukami et al.~\cite{fukami2021sparse} introduced \ac{SINDy}, which identifies sparse representations of the underlying dynamics from low-dimensionalized flow data, offering a different perspective on capturing the temporal evolution of fluid flows.

Moreover, \ac{ConvLSTM}, an end-to-end spatio-temporal deep learning prediction method, effectively integrates convolutional structures with LSTM networks~\cite{shi2015convolutional}, which can process spatial information through convolutional layers and temporal sequences via \ac{LSTM} units. For instance, Huang et al.~\cite{huang2022predictions} demonstrated the application of ConvLSTM in the prediction of flow and temperature fields in a T junction, showing its ability to handle complex dynamic systems. Similarly, Beiki and Kamali~\cite{beiki2023novel} proposed a novel attention-based \ac{CAE} and ConvLSTM for reduced-order modeling in fluid mechanics, highlighting its effectiveness in processing spatio-temporal data.

However, purely data-driven methods often struggle with ensuring generalization and providing physically realistic outputs during rolling forecasts with limited training data~\cite{wu2023deep}. Recently, physics constrained neural network has been used to improve the accuracy and generalizability of the spatio-temporal model for dynamical systems with explicit formulas by introducing physics constraints during the training process~\cite{xie2022physics, gong2022data, zhou2024multi}. For instance, Ouala et al. (2023)~\cite{ouala2023bounded} demonstrated how physics-constrained models, particularly through neural ordinary differential equation models with linear-quadratic dynamics and global boundedness constraints, can effectively address the challenges of spatio-temporal prediction in partially observed dynamical systems. Recent work by Zhou et al.~\cite{zhou2024multi} demonstrated an application of CAE-LSTM with multi-fidelity physics constraints, effectively tackling the complexities of training and data fidelity in dynamical systems prediction. 

Despite the advantages of purely data-driven convolutional network-based methods and physics-based approaches previously mentioned, as outlined in works such as Maturana and Scherer (2015)~\cite{maturana2015voxnet} and Liu et al. (2016)~\cite{liu2016lasagna}, they both meet difficulties when processing unstructured data and movable sensors. Applications in real-world problems often involve sparse and time-varying observations, which do not align well with the structural requirements of convolutional layers that are designed to operate on uniformly structured and spatially consistent data~\cite{fukami2019super,eberendu2016unstructured, park2016unstructured}. 

Given the sparse nature of dynamical systems, \ac{GNNs} are applied to handle unstructured data due to their ability to model complex connectivity~\cite{zhou2020fully,shi2022gnn}. Meanwhile, the GNN-LSTM framework is utilized for spatio-temporal prediction in nonlinear dynamical systems with sparse observations~\cite{kuo2024gnn,gong2021mmpoint,huang2023deep}. Although these methods excel in processing unstructured data, they mainly capture temporal and spatial dependencies within densely spatial data in unstructured data~\cite{li2017diffusion,yu2017spatio} and require significant computational resources during training and inference~\cite{shi2022gnn}. 

For dealing with issues of sparse and unstructured observations in general dynamical systems, Fukami et al. (2021)~\cite{fukami2021global} also utilized Voronoi tessellation to interpolate sparse and time-varying data in CNN-based field reconstruction. Meanwhile, traditional Kriging methods, such as Kriging combined with regression for spatio-temporal forecasting~\cite{zhu2010comparing, yang2013solar} and \ac{3D-Kriging}, which considers time as an additional input, are also used to predict future states~\cite{xiao2020improved, li2012interpolation} but are highly sensitive to parameter settings, require high computational resources, and struggle with nonlinear prediction for non-Gaussian distributions~\cite{myers2000improving, erdogan2022combination,chen2020deepkriging}. Although Voronoi tessellation reduces computational demands and is less sensitive to parameters, it is important to note that while Fukami et al. (2021)~\cite{fukami2021global} used it to make significant strides in field reconstruction, they did not address sequence-to-sequence prediction, which involves more complex dynamics such as those with movable sensors. Furthermore, the lack of physics constraints in their approach may lead to ill-defined problems in sparse prediction scenarios, potentially resulting in infinitely many solutions~\cite{brunton2020machine}.

To address the challenges, we propose a novel framework named \ac{DSOVT}, which consists of two algorithms: \ac{CED-LSTM} and \ac{ConvLSTM}. This framework utilizes Voronoi tessellation to interpolate the fields of dynamical systems from sparse observations~\cite{aurenhammer2000voronoi}, which efficiently constructs a structured grid representation, providing a homogeneous basis for subsequent predictive process. For our \ac{CED-LSTM} in \ac{DSOVT} framework, we firstly use \ac{CED} to extract the latent representation of Voronoi tessellations. This \ac{CED} consists of two key stages: first, an encoder transforms the Voronoi tessellations into compact, low-dimensional latent spaces; then, a decoder maps these latent representations back to true state fields.
We then constructed an \ac{LSTM} model for temporal prediction in the latent space. Unlike the \ac{CED-LSTM} in the \ac{DSOVT} framework, the \ac{ConvLSTM} model is a more streamlined, end-to-end system. It takes Voronoi tessellations as inputs and directly generates true state fields as outputs, specifically for spatio-temporal predictions. This approach effectively reduces the model's complexity by eliminating the need for intermediate latent representations. However, due to its larger number of parameters, which can lead to higher model capacity, the \ac{ConvLSTM} may require more data for training to effectively generalize and achieve optimal performance.

In this paper, we integrate physics constraint losses with traditional data-driven loss functions, such as \ac{MSE} loss, to enhance the training process of the \ac{DSOVT} framework. For the \ac{CED-LSTM}, physics-constrained loss functions are incorporated into \ac{LSTM} training. In the \ac{ConvLSTM}, this combined loss approach directly governs the end-to-end training process. This method effectively bridges data-driven insights with physical laws, improving the robustness and reliability of predictions.

As proof of concept, we performed numerical experiments on \ac{NOAA SST} data~\cite{huang2020noaa} and shallow water systems with explicit formulas~\cite{de1871theorie}. We compared our framework against traditional methods such as \ac{2D-Kriging}~\cite{van2020spatio} and \ac{3D-Kriging}~\cite{snepvangers2003soil}. \ac{2D-Kriging} is a geostatistical interpolation technique that predicts environmental variables at unsampled locations in two dimensions with linear regression, while \ac{3D-Kriging} extends this approach to three dimensions, adding time to provide spatio-temporal interpolation capabilities.
For the evaluation of prediction quality, three distinct metrics were utilized: the \ac{PSNR}, the \ac{SSIM}, and \ac{R-RMSE}. The \ac{PSNR} is primarily focused on quantifying errors at the pixel level, whereas the \ac{SSIM} is employed to assess the overall similarity between the predicted images and the actual reality of the ground~\cite{sara2019image}. \ac{R-RMSE} is used to evaluate the relative magnitude of errors across different scales, providing a normalized measure of error intensity~\cite{willmott2005advantages}. They allow us to fully assess the accuracy of the model predictions from both the local detail and the overall coherence perspectives~\cite{sara2019image}. To evaluate computational efficacy, we also calculated the inference time for different methods.

In summary, we make the following main contributions in this study,
\begin{itemize}
    \item We introduce a novel spatio-temporal prediction framework for dynamical systems based on Voronoi tessellation, named \ac{DSOVT}. This framework is capable of handling and predicting data from sparse and time-varying sensors. For dynamical systems with explicit formulas, we integrate physics constraints into our \ac{DSOVT} model, demonstrating improvements in rolling forecast accuracy.
    
    \item \ac{DSOVT} surpasses methods like \ac{2D-Kriging} and \ac{3D-Kriging} in spatio-temporal forecasting during multi-step predictions. In tests on the \ac{NOAA SST} dataset, \ac{CED-LSTM} outperforms \ac{2D-Kriging} with a 37.70\% increase in \ac{SSIM}, a 44.90\% increase in \ac{PSNR}, and a 77.14\% increase in \ac{R-RMSE}, while halving the inference times. \ac{ConvLSTM} achieves improvements of 22.95\% in \ac{SSIM}, 17.8\% in \ac{PSNR}, and 62.86\% in \ac{R-RMSE}. For shallow water systems, compared to \ac{2D-Kriging}, \ac{CED-LSTM} achieves improvements of 20.83\% in \ac{SSIM}, 56.21\% in \ac{PSNR}, and 81.48\% in \ac{R-RMSE}, with a 95\% reduction in inference time. Additionally, \ac{ConvLSTM} improves \ac{SSIM} by 20.83\%, \ac{PSNR} by 56\%, and \ac{R-RMSE} by 74.07\%, cutting inference time from 17.44 seconds to 6.31 seconds.

    \item In shallow water systems, integrating physics constraints into \ac{DSOVT} enhances its rolling forecast accuracy, even with limited data availability. Compared to purely data-driven approaches, physics-constrained \ac{CED-LSTM} model shows improvements of 5.44\% in \ac{SSIM}, 4.84\% in \ac{PSNR}, and 22.22\% in \ac{R-RMSE}. Similarly, \ac{ConvLSTM} increases \ac{SSIM} by 20.96\%, \ac{PSNR} by 11.53\%, and \ac{R-RMSE} by 26.15\%. These physics-constrained models stabilize and refine predictions, significantly reducing large errors.
\end{itemize}

 The structure of this paper is outlined as follows: Section~\ref{sec:methodology} delves into the methodology of our framework for spatio-temporal prediction for sparse fields. Numerical experiments involving \ac{NOAA SST} data and shallow water system are explored in Section~\ref{sec:experiments-NOAA} and~\ref{sec:experiments-SW}. The paper ends in Section~\ref{sec:Conclusion} with a synthesis of the key insights gathered and future work.

\section{Methodology}
\label{sec:methodology}
The proposed \ac{DSOVT} comprises two algorithms: one based on the \ac{CED} architecture combined with \ac{LSTM}, and another utilizing end-to-end \ac{ConvLSTM}. For the problem of sparse and time-varying sensors, we design and optimize our framework based on Voronoi tessellation and physics constraints.
\begin{figure*}[h!]
\centering
\includegraphics[width=0.8\textwidth]{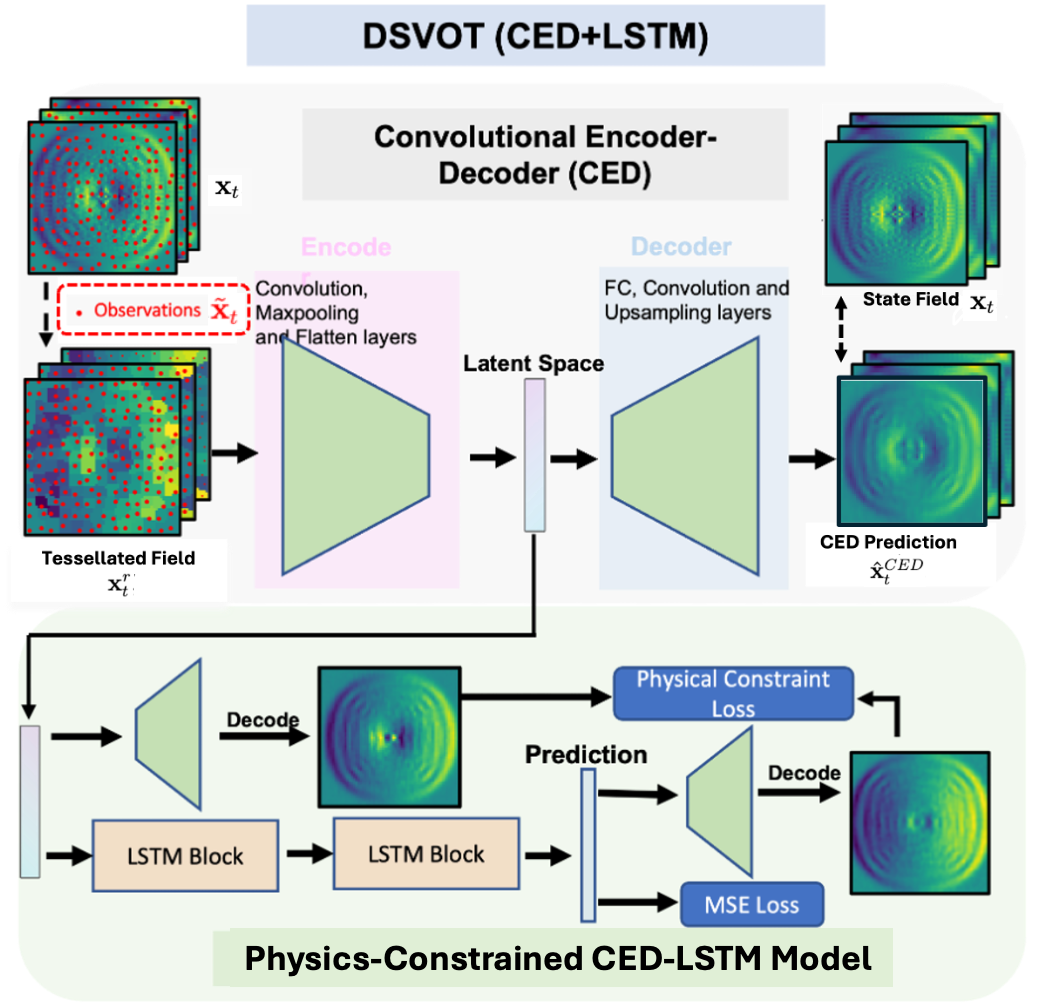}
\caption{Schematic representation of physics-constrained CED-LSTM model employing Voronoi tessellation for enhanced state field mapping from sparse observations.}
\label{fig:phys-constr-CED-LSTM-Voronoi}
\end{figure*}

\subsection{Voronoi Tessellation for Sparse Interpolation}

Sparse and time-varying observations across the global domain are common in dynamical systems, necessitating the selection of an effective unstructured interpolation method. Fukami et al. (2021) proposed a \ac{VCNN} approach, demonstrating the efficacy in utilising Voronoi tessellation to address the challenges posed by sparse and unstructured data~\cite{fukami2021global}.

In our study, we define a three-dimensional state field at time $t$, denoted by $\mathbf{x}_t \in \mathbb{R}^{N_x \times N_y \times N_c}$. Here, $N_x$ and $N_y$ represent spatial dimensions and $N_c$ denotes the number of channels. Given that Voronoi tessellation is fundamentally a 2D interpolation technique, and considering that our field includes a third dimension represented by channel dimensions \(N_c\), we extend the method to handle multi-channel data by applying tessellation independently to each channel. To simplify the explanation of the Voronoi tessellation process, we consider the case where \(N_c = 1\), implying that each channel is treated separately but identically.

As illustrated in Figure~\ref{fig:phys-constr-CED-LSTM-Voronoi}, on the state field $\mathbf{x}_t$, there are $K$ local sensors (red dots) at locations $\{(i_{t,k}, j_{t,k})\} \in [1, N_x] \times [1, N_y]$ for the $k^{th}$ sensor at time $t$, where $K < N_x \times N_y$. Sensors are randomly distributed across the state field $\mathbf{x}_t$. Notably, the $k^{th}$ sensor locations $\{(i_{t,k}, j_{t,k})\}$ remain consistent across all channels and may vary over time $t$. Each observed value at these sensor locations, \({o}_{t,k}\), is obtained from the state field as:
\begin{equation}
    o_{t,k} = \mathbf{x}_t(i_{t,k}, j_{t,k})
\end{equation}

We aim to interpolate the observation domain $\tilde{\mathbf{x}}_t = \{o_{t,k} \mid k = 1, \dots, K\} \in \mathbb{R}^{2}$ from the set of sparse and limited sensor observations $\{(i_{t,k}, j_{t,k}) \mid k = 1, \dots, K\}$. A Voronoi cell \(R_{t,k}\) associated with the $k^{th}$ sensor, located at coordinates $(i_{t,k}, j_{t,k})$ and with observation values $o_{t,k}$, can be defined as:
\begin{equation}
R_{t,k} = \left\{ (i_x, j_x) \mid d\left((i_x, j_x), (i_{t,k}, j_{t,k})\right) \leq d\left((i_x, j_x), (i_{t,q}, j_{t,q})\right) \text{ for all } q \neq k \text{ and } 1 \leq q \leq K \right\}
\end{equation}
where $d(\cdot)$ denotes the Euclidean distance between two points.

Therefore, the given observation domain can be partitioned into several Voronoi cells $\{R_{t,k} \mid k = 1, \ldots, K\}$. This partitioning can be mathematically described as follows:
\begin{equation}
\left\{(i_x, j_x)\right\} = \bigcup_{k=1}^{K} R_{t,k} \quad \text{with} \quad R_{t,k} \cap R_{t,q} = \emptyset \quad \text{for all } k \neq q.
\end{equation}

For each region $R_{t,k}$, we define $\tilde{o}_{t, R_{t,k}}$ as the assigned value at time $t$ for the region $R_{t,k}$, using the observation value $o_{t,k}$:
\begin{equation}
\tilde{o}_{t, R_{t,k}} = o_{t,k} \quad \text{for } k = 1, \ldots, K.
\label{eq:obs_vor}
\end{equation}

As a result, a tessellated observation field $\mathbf{x}^r_t = \{ \tilde{o}_{t, R_{t,k}} \mid k = 1, \ldots, K\} \in \mathbb{R}^{N_x \times N_y}$ is created.
In our study, we replicate the tessellation process independently across the $N_c$ channels, resulting in a tessellated three-dimensional field $\mathbf{x}^r_t \in \mathbb{R}^{N_x \times N_y \times N_c}$. This interpolated field forms the foundation for subsequent spatio-temporal predictions.

\subsection{Latent Representation Extraction from Voronoi Tessellation}

In our study, we deploy a CED (encoder-decoder) architecture specifically designed for processing Voronoi tessellation. This architecture excels at extracting features from sparse representations of Voronoi fields $\mathbf{x}^r_t$ and effectively reconstructing the latent representations back into the state fields $\mathbf{x}_t$. The encoder, \( \mathscr{F}_e \), converts these fields into a compact latent space representation, $\mathbf{h}_t$, which reduces data dimensionality while preserving essential information. Crucially, this compact representation also reduces the computational resources required for subsequent time series prediction, thereby enhancing the efficiency and effectiveness of predictive modelling. The decoder \( \mathscr{F}_d \), is then used to reconstruct \( \mathbf{h}_t \) back into the CED reconstructed state fields \( \hat{\mathbf{x}}^{CED}_{t} \). This step is vital for precisely restoring the full data structure of the state fields and for replicating the dynamics inherent in the original Voronoi tessellations. Efficient data compression and subsequent reconstruction using the CED architecture significantly reduce computational burden during processing, enhancing operational efficiency. This dual functionality is described by the following equations:

\begin{equation}
\mathbf{h}_t = \mathscr{F}_e(\mathbf{x}^r_t) \quad \text{and} \quad \hat{\mathbf{x}}^{CED}_{t} = \mathscr{F}_d(\mathbf{h}_t)
\label{eq:process_ced}
\end{equation}

The state field $\mathbf{x}_t$ represents the full field data, which is known across the entire domain in our training dataset. The observation domain is represented by \(\tilde{\mathbf{x}}_t = \{o_{t,k} \mid k = 1, \dots, K\} \in \mathbb{R}^2\), which is derived from a set of sparse and limited sensor observations. Specifically, the sensors capture data from the state fields $\mathbf{x}_t$. The field derived from Voronoi tessellation, denoted by \(\mathbf{x}_t^r\), represents the interpolated data from these sensors. The primary function of our CED architecture is to learn an efficient transformation from these sparse and potentially noisy inputs ($\mathbf{x}_t^r$) back to the full-field state ($\hat{\mathbf{x}}^{CED}_{t}$), which approximates the true state field $\mathbf{x}_t$.

The training of the CED involves minimizing the reconstruction error by approximating the $\hat{\mathbf{x}}^{CED}_{t}$ with state fields $\mathbf{x}_t$. The MSE serves as the commonly used loss function, given by:
\begin{equation}
\mathcal{L}_{\text{CED}}(\theta_{e}, \theta_{d}) = \frac{1}{T} \sum_{t=1}^{T} \|\mathbf{x}_{t} - \hat{\mathbf{x}}^{CED}_{t}\|^2
\end{equation}

$T$ indicates the total number of time steps, and $\theta_{e}$ and $\theta_{d}$ are the parameters of the encoder and decoder in the CED, respectively. The term $\frac{1}{T} \sum_{t=1}^{T} \|\mathbf{x}_{t} - \hat{\mathbf{x}}^{CED}_{t}\|^2$ quantifies the MSE between the state fields $\mathbf{x}_t$ and the reconstructed fields $\hat{\mathbf{x}}^{CED}_{t}$ for $t=1,2,\ldots,T$.

The architecture of CED is described in Table~\ref{table: CED_structure}, illustrating the flow from input $\mathbf{x}^r_t$ to output $\mathbf{x}_t$ through the \ac{CED} framework.
The custom activation function mentioned in Table~\ref{table: CED_structure} is designed to meet the specific requirements of the CED architecture for field reconstruction.

\begin{table}[h!]
\centering
\caption{Structure of the \ac{CED} with $\mathbf{x}^r_t$ as input and $\mathbf{x}_t$ as output. $Z$ represents the size of the latent space.}
\begin{tabular}{ccc} 
    \toprule
    \textbf{Layer (type)} & \textbf{Output Shape} & \textbf{Activation} \\ 
    \midrule
    Input  & $(N_x, N_y, N_c)$ & \\
    Conv2D $(3\times3)$  & $(N_x, N_y, 32)$ & ReLU \\
    MaxPooling2D $(2\times2)$  & $\left(\frac{N_x}{2}, \frac{N_y}{2}, 32\right)$ & \\
    Conv2D $(3\times3)$  & $\left(\frac{N_x}{2}, \frac{N_y}{2}, 64\right)$ & ReLU \\
    MaxPooling2D $(2\times2)$  & $\left(\frac{N_x}{4}, \frac{N_y}{4}, 64\right)$ & \\
    Conv2D $(3\times3)$  & $\left(\frac{N_x}{4}, \frac{N_y}{4}, 128\right)$ & ReLU \\
    MaxPooling2D $(2\times2)$  & $\left(\frac{N_x}{8}, \frac{N_y}{8}, 128\right)$ & \\  
    Flatten  & $\left(\frac{N_x}{8} \times \frac{N_y}{8} \times 128\right)$  & \\ 
    Dense  & $(Z)$ & ReLU \\ 
    Dense  & $\left(\frac{N_x}{8} \times \frac{N_y}{8} \times 128\right)$  & ReLU \\
    Reshape  & $\left(\frac{N_x}{8}, \frac{N_y}{8}, 128\right)$ & \\
    Conv2D $(3\times3)$  & $\left(\frac{N_x}{8}, \frac{N_y}{8}, 128\right)$ & ReLU \\
    UpSampling2D $(2\times2)$  & $\left(\frac{N_x}{4}, \frac{N_y}{4}, 128\right)$ & \\
    Conv2D $(3\times3)$  & $\left(\frac{N_x}{4}, \frac{N_y}{4}, 64\right)$ & ReLU \\
    UpSampling2D $(2\times2)$  & $\left(\frac{N_x}{2}, \frac{N_y}{2}, 64\right)$ & \\
    Conv2D $(3\times3)$  & $\left(\frac{N_x}{2}, \frac{N_y}{2}, 32\right)$ & ReLU \\
    UpSampling2D $(2\times2)$  & $(N_x, N_y, 32)$ & \\
    Conv2D $(3\times3)$  & $(N_x, N_y, N_c)$ & Custom Activation \\
    \bottomrule
\end{tabular}
\label{table: CED_structure}
\end{table}

The \ac{CED} component, illustrated in Figure~\ref{fig:phys-constr-CED-LSTM-Voronoi} and detailed in Algorithm~\ref{algo:CED}, exemplifies how to extract latent representations from Voronoi tessellation fields and reconstruct them into state fields.
\begin{algorithm}
\caption{Latent Representation Extraction with CED}
\label{algo:CED}
\begin{algorithmic}[1]
\State Inputs: $\mathbf{x}_t$ (state fields), $\tilde{\mathbf{x}}_t$ (observation fields).
\State Parameters: $N_{\text{epoch}}$ (number of training epochs), $\eta$ (learning rate), $\theta_{e}$ (encoder parameters), $\theta_{d}$ (decoder parameters),$\mathcal{L}_{CED}$ (loss function of CED).
\State $\mathbf{x}^r_t$ $\xleftarrow{\text{Voronoi}}$ $\tilde{\mathbf{x}}_t$
\For{$n = 1$ to $N_{\text{epoch}}$}
    \State \textbf{Encoding:} $\mathbf{h}_t = \mathscr{F}_e(\mathbf{x}^r_{t})$ 
    \State \textbf{Decoding:} $\hat{\mathbf{x}}^{CED}_{t} = \mathscr{F}_d(\mathbf{h}_t)$ 
    \State \textbf{Compute Loss:} $\mathcal{L}_{CED,n} = \|\mathbf{x}_{t} - \hat{\mathbf{x}}^{CED}_{t}\|^2$
    \State \textbf{Update Parameters:} Backpropagate to compute $\nabla_{\theta_{e}, \theta_{d}} \mathcal{L}_{CED,n}$ and update $\theta_{e}, \theta_{d}$ with Adam using learning rate $\eta$
\EndFor
\end{algorithmic}
\end{algorithm}

\subsection{LSTM with latent representation}
After extracting the latent spatial representation $\mathbf{h}_t$ from \ac{CED}, our methodology uses \ac{LSTM} to learn the dynamics of the system. The \ac{LSTM}'s ability to handle sequence-to-sequence tasks makes it a powerful tool for predicting future states of physical systems based on past observations encoded in latent space~\cite{yildirim2019new,zhao2019time}.

Given the total sequence length \( T \), the input sequence length \( S_{\text{in}} \), and the output sequence length \( S_{\text{out}} \), an LSTM network is designed to predict a sequence of future latent spatial representations from an input sequence of latent representations. Specifically, the network predicts the sequence \( \{\mathbf{h}_t \mid t = i+S_{\text{in}}, \dots, i+S_{\text{in}}+S_{\text{out}}-1\} \), based on the input sequence \( \{\mathbf{h}_t \mid t = i, \dots, i+S_{\text{in}}-1\} \). The predictions are denoted as \( \{\hat{\mathbf{h}}_t \mid t = i+S_{\text{in}}, \dots, i+S_{\text{in}}+S_{\text{out}}-1\} \). The index \( i \) serves as the starting index of the input sequence, ranging from 1 to \( T-S_{\text{in}}-S_{\text{out}}+1 \). This indexing method enables the LSTM to perform multi-step predictions by sequentially shifting the input data window across the whole dataset.
\begin{align}
\text{Input sequence:} & \quad \left[\mathbf{h}_i, \ldots, \mathbf{h}_{i+S_{\text{in}}-1}\right] \\
\text{Model trains to predict the output sequence:} & \quad \left[\mathbf{h}_{i+S_{\text{in}}}, \ldots, \mathbf{h}_{i+S_{\text{in}}+S_{\text{out}}-1}\right] \\
\text{Where the predicted sequence is:} & \quad \left[\hat{\mathbf{h}}_{i+S_{\text{in}}}, \ldots, \hat{\mathbf{h}}_{i+S_{\text{in}}+S_{\text{out}}-1}\right]
\end{align}

The LSTM utilizes the \ac{MSE} as the loss function to minimize the difference between the predicted outputs and the corresponding true latent representations. The \ac{MSE} is given by:

\begin{align}
\text{MSE} = \frac{1}{S_{\text{out}}} \sum_{j=0}^{S_{\text{out}}-1} \left( \mathbf{h}_{i+S_{\text{in}}+j} - \hat{\mathbf{h}}_{i+S_{\text{in}}+j} \right)^2
\end{align}

\subsection{ConvLSTM with Voronoi Tessellation Inputs}

\begin{figure*}[h!]
\centering
\includegraphics[width=1\textwidth]{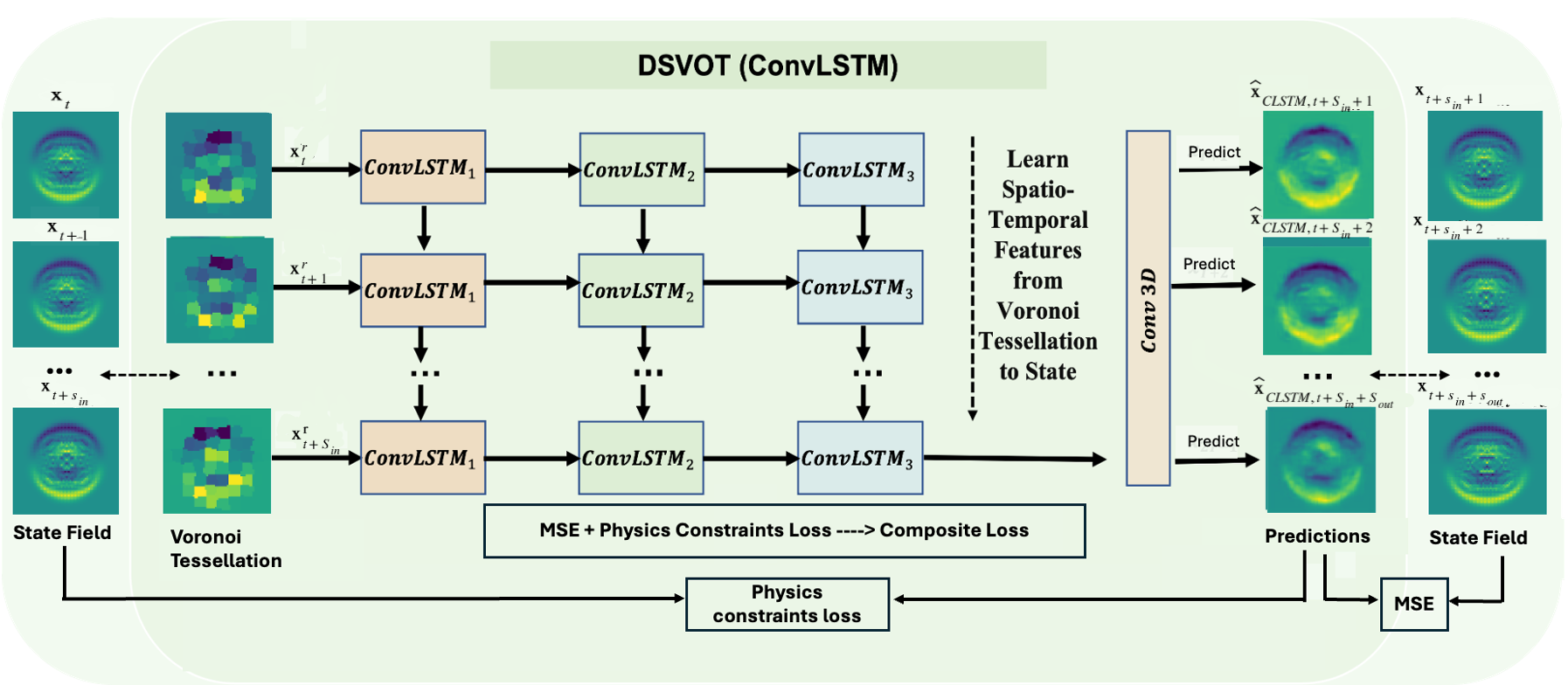}
\caption{Schematic representation of physics-constrained ConvLSTM model employing Voronoi tessellation to capture spatial dependencies and predict future state fields in dynamical systems. The process starts at time $t$, using $S_{in}$ and $S_{out}$ as the lengths of the input and output sequences, respectively.} 
\label{fig:vscf_model}
\end{figure*}

As shown in Figure~\ref{fig:vscf_model}, the \ac{ConvLSTM} model, processing Voronoi tessellation inputs, uses convolutional operations within its \ac{LSTM} units to effectively handle spatial dependencies. This integrated approach not only preserves the temporal sequence modelling capabilities inherent to traditional \ac{LSTM} but also leverages the proficiency of the convolutional network in processing spatial information. The end-to-end architecture of \ac{ConvLSTM} makes it a highly effective tool for predicting spatio-temporal dynamical systems directly related to the state fields.

Given a field starting at index \( i \), where \( i \in \{1, \ldots, T-S_{in}-S_{\text{out}}+1\} \), the \ac{ConvLSTM} model processes the tessellation inputs, $\{\mathbf{x}^r_t \mid t = i, \ldots, i+S_{\text{in}}-1\}$, with the aim of predicting the future state sequence $\{\mathbf{x}_t \mid t = i+S_{\text{in}}, \ldots, i+S_{\text{in}}+S_{\text{out}}-1\}$, capturing the evolution of spatial dynamics over time. The training and multi-step prediction process is formalized as follows:
\begin{align}
\text{Input fields:} & \quad \left[\mathbf{x}^r_i, \ldots, \mathbf{x}^r_{i+S_{\text{in}}-1}\right] \\
\text{The model trains to predict the output fields:} & \quad \left[\mathbf{x}_{i+S_{\text{in}}}, \ldots, \mathbf{x}_{i+S_{\text{in}}+S_{\text{out}}-1}\right] \\
\text{Where the prediction sequence is:} & \quad \left[\hat{\mathbf{x}}^{CLSTM}_{i+S_{\text{in}}}, \ldots, \hat{\mathbf{x}}^{CLSTM}_{i+S_{\text{in}}+S_{\text{out}}-1}\right]
\end{align}

Here, $\hat{\mathbf{x}}^{CLSTM}_{t}$ denotes the \ac{ConvLSTM}'s prediction at time \(t\). We continue to use \ac{MSE} as our loss function, which is calculated as follows:
\begin{align}
\text{MSE} = \frac{1}{S_{\text{out}}} \sum_{j=0}^{S_{\text{out}}-1} \left( \mathbf{x}_{i+S_{\text{in}}+j} - \hat{\mathbf{x}}^{CLSTM}_{i+S_{\text{in}}+j} \right)^2
\end{align}

\subsection{Physics Constraints}
In this section, we focus on the application of energy conservation principles within dynamical systems with explicit formulas. Energy conservation is a fundamental principle in physical simulations, particularly in fluid dynamics and thermal processes~\cite{norton2013computational}. Compared to purely data-driven \ac{ML} models, integrating these principles helps ensure that models not only predict more realistic outcomes but also adhere to the fundamental laws of physics.The works of Zhou et al. (2024)~\cite{zhou2024multi} and Guo et al. (2020)~\cite{guo2020solving} serve as examples, illustrating how incorporating physics constraints can significantly enhance the accuracy and reliability of spatio-temporal prediction for dynamical systems. Through this approach, we aim to bridge the gap between computational physics and deep learning models, ensuring that our models are both scientifically rigorous and practically viable.

Energy conservation asserts that in a conservative system, the total energy remains constant over time. This concept is particularly relevant in systems where external energy exchanges are absent.
To quantify alignment with energy conservation principles, we define an energy conservation loss function, \(\mathcal{L}_{\text{energy}}\), which measures the discrepancy between the energy states of the input and output fields. This function is integrated into the overall loss function to enhance the adherence of the model to energy conservation. Assuming that \(S_{\text{in}}\) is equal to \(S_{\text{out}}\), the objective in this conservative model framework is to ensure that the energy of the output fields, $\mathbf{e}_{\text{out}}$, matches that of the input fields, $\mathbf{e}_{\text{in}}$.

The formulation of \(\mathcal{L}_{\text{energy}}\) is as follows:
\begin{flalign}
    &\mathbf{e}_{\text{in}} = \frac{1}{S_{\text{in}}} \sum_{i=t}^{t+S_{\text{in}}-1} \mathcal{E}(\mathbf{x}_{i})&\\
    &\mathbf{e}_{\text{out}} = \begin{cases} 
        \frac{1}{S_{\text{out}}} \sum_{i=t+S_{\text{in}}}^{t+S_{\text{in}}+S_{\text{out}}-1} \mathcal{E}(\mathscr{F}_d(\hat{\mathbf{h}}_i)), & \text{for CED-LSTM} \\
        \frac{1}{S_{\text{out}}} \sum_{i=t+S_{\text{in}}}^{t+S_{\text{in}}+S_{\text{out}}-1} \mathcal{E}(\hat{\mathbf{x}}^{CLSTM}_{i}), & \text{for ConvLSTM}
    \end{cases} &\\
    &\mathcal{L}_{\text{energy}} = |\mathbf{e}_{\text{in}} - \mathbf{e}_{\text{out}}| &
    \label{loss_energy}
\end{flalign}

Here, \(\mathcal{E}\) represents the calculation of energy, and \(| \cdot |\) denotes the absolute value.

\subsubsection{Incorporating Energy Conservation Constraints in CED-LSTM Training}

To enhance the accuracy and physical consistency of our model, we integrate energy conservation constraints into the LSTM training process. Initially, the LSTM network leverages current parameters to predict latent spatial representations. These predictions, denoted as $\{\hat{\mathbf{h}}_t \mid t=1, \ldots, T\}$, are subsequently decoded into predicted fields $\{\hat{\mathbf{x}}^{LSTM}_{t} \mid t=1, \ldots, T\}$, which are equivalent to $\mathscr{F}_d(\{\hat{\mathbf{h}}_t \mid t=1, \ldots, T\})$. These predictions are compared with the state fields $\{\mathbf{x}_t \mid t=1, \ldots, T\}$ to compute the energy conservation loss $\mathcal{L}_{\text{energy}}$ (as defined in Equation~\ref{loss_energy}), in addition to the \ac{MSE} loss. To incorporate this into the training of CED-LSTM models, we introduce a weighting coefficient for the energy conservation loss, \(\lambda_{\text{energy}}\), which leading to a modified composite loss function \(\mathcal{L}_{total}\):
\begin{equation}
    \mathcal{L}_{total} =\frac{1}{S_{\text{out}}} \sum_{i=t+S_{\text{in}}}^{t+S_{\text{in}}+S_{\text{out}}-1} \|\mathbf{h}_{i} - \hat{\mathbf{h}}_{i}\|^2 + \lambda_{\text{energy}} \cdot \mathcal{L}_{\text{energy}}
    \label{eq:loss_func_phy_ced_lstm}
\end{equation}
This approach ensures that the predictions conform to physical laws, as detailed in Algorithm~\ref{algo: LSTM-PHY}.

Physics-constrained CED-LSTM models employ a rolling forecast strategy to get long-term predictions. This approach leverages a continuous feedback loop where each output sequence is immediately used as the input for the next cycle to get the predictions. By characterizing the error accumulation of rolling forecasts, CED-LSTM provides a clear view of how physics constraints influence its performance, enhancing our understanding of error dynamics and overall model stability. The starting index \(i\) for predictions ensures sufficient data for both input and output sequences, governed by the condition:
\[
1 \leq i \leq T - S_{\text{in}} - S_{\text{out}} + 1
\]

The process of rolling forecast in CED-LSTM is then expressed as follows:
\begin{align}
\label{rolling_phy_cedlstm}
\text{Initial prediction:} & \quad \left[\mathbf{h}_i, \ldots, \mathbf{h}_{i+S_{\text{in}}-1}\right] \xrightarrow{\text{Prediction}} \left[\hat{\mathbf{h}}_{i+S_{\text{in}}}, \ldots, \hat{\mathbf{h}}_{i+S_{\text{in}}+S_{\text{out}}-1}\right]\\
\label{rolling_phy_cedlstm_2}
\text{Rolling forecast:} & \quad \left[\hat{\mathbf{h}}_{i+S_{\text{in}}}, \ldots, \hat{\mathbf{h}}_{i+S_{\text{in}}+S_{\text{out}}-1}\right] \xrightarrow{\text{Prediction}} \left[\hat{\mathbf{h}}_{i+S_{\text{in}}+S_{\text{out}}}, \ldots, \hat{\mathbf{h}}_{i+S_{\text{in}}+2S_{\text{out}}-1}\right]
\end{align}

\begin{algorithm}
\caption{Training a physics-constrained CED-LSTM}
\label{algo: LSTM-PHY}
\begin{algorithmic}[1]

\State \textbf{Inputs:} $\mathbf{x}_t$ (state fields), $\mathbf{x}^r_t$ (tessellated observation fields).
\State \textbf{Initial CED Processing:} Train CED from $\mathbf{x}^r_t$ to $\mathbf{x}_t$ according to Algorithm~\ref{algo:CED} to obtain the pre-trained decoder $\mathscr{F}_d$.
\State \textbf{Parameters:} $N_{\text{epoch}}$ (total epochs), $\eta$ (learning rate), $\theta_{\text{LSTM}}$ (LSTM model parameters), $T$ (total timesteps), $S_{\text{in}}$ (input sequence length), $S_{\text{out}}$ (output sequence length), $\lambda_{\text{energy}}$ (energy conservation loss weight), $\mathcal{L}_{\text{total}}$ (composite loss function), $\mathcal{E}$ (energy calculation function), $\mathscr{F}_d$ (decoder of CED).

\State $\mathbf{h}_t = \mathscr{F}_d(\mathbf{x}^r_t)$
\For{$n = 1$ to $N_{\text{epoch}}$}
    \For{$t = 1$ to $T - S_{\text{in}} - S_{\text{out}} + 1$}
        \State \textbf{Forward Pass via LSTM model:}
        \State $\hat{\mathbf{h}}_{t+S_{\text{in}}:t+S_{\text{in}}+S_{\text{out}}-1} = \text{LSTM}(\mathbf{h}_{t:t+S_{\text{in}}-1}; \theta_{\text{LSTM}})$
        
        \State \textbf{Decode Predicted States:}
        \State $\hat{\mathbf{x}}^{LSTM}_{ t+S_{\text{in}}:t+S_{\text{in}}+S_{\text{out}}-1} = \mathscr{F}_d(\hat{\mathbf{h}}_{t+S_{\text{in}}:t+S_{\text{in}}+S_{\text{out}}-1})$

        \State \textbf{Compute Physics Constraint Loss:}
        \State $\mathcal{L}_{\text{energy}} = \left| \mathcal{E}(\hat{\mathbf{x}}^{LSTM}_{ t+S_{\text{in}}:t+S_{\text{in}}+S_{\text{out}}-1}) - \mathcal{E}(\mathbf{x}_{t:t+S_{\text{in}}-1}) \right|$

        \State \textbf{Compute Composite Loss:}
        \State $\mathcal{L}_{\text{total}} = \|\hat{\mathbf{h}}_{t+S_{\text{in}}:t+S_{\text{in}}+S_{\text{out}}-1} - \mathbf{h}_{t+S_{\text{in}}:t+S_{\text{in}}+S_{\text{out}}-1}\|^2 + \lambda_{\text{energy}} \cdot \mathcal{L}_{\text{energy}}$

        \State \textbf{Update Parameters:}
        \State Backpropagate to compute $\nabla_{\theta_{\text{LSTM}}} \mathcal{L}_{\text{total}}$ and update $\theta_{\text{LSTM}}$ using the Adam optimizer with $\eta$
    \EndFor
\EndFor

\end{algorithmic}
\end{algorithm}

\subsubsection{Physics-constrained ConvLSTM for Spatio-Temporal Prediction}

Building on the principles discussed earlier, we now explore the specifics of applying physics constraints within \ac{ConvLSTM} models for spatio-temporal predictions. In the physics-constrained training process, \ac{ConvLSTM} integrates a loss computation step that combines both the \ac{MSE} and energy conservation constraints, as detailed in Equation~\ref{eq:loss_func_phy_convlstm} and illustrated in the loss computation section of Figure~\ref{fig:vscf_model}. The model parameters are updated on the basis of this composite loss, with the dual objective of minimising prediction error and ensuring adherence to physical laws in dynamical systems. This dual focus on prediction accuracy and physical consistency is central to our \ac{ConvLSTM} training process, as outlined in Algorithm~\ref{algo: Physics-ConvLSTM}.
\begin{equation}
    \mathcal{L}_{total} =
        \frac{1}{S_{\text{out}}} \sum_{i=t+S_{\text{in}}}^{t+S_{\text{in}}+S_{\text{out}}-1} \|\mathbf{x}_{i} - \hat{\mathbf{x}}^{CLSTM}_{i}\|^2 + \lambda_{\text{energy}} \cdot \mathcal{L}_{\text{energy}}.
    \label{eq:loss_func_phy_convlstm}
\end{equation}
For making predictions, we employ a rolling forecast strategy:
\begin{align}
\text{Initial prediction:} & \quad \left[\mathbf{x}^r_i, \ldots, \mathbf{x}^r_{i+S_{\text{in}}-1}\right] \xrightarrow{\text{Prediction}} \nonumber \\
& \quad \left[\hat{\mathbf{x}}^{CLSTM}_{i+S_{\text{in}}}, \ldots, \hat{\mathbf{x}}^{CLSTM}_{i+S_{\text{in}}+S_{\text{out}}-1}\right] \label{eq:initial_prediction} \\
\text{Rolling forecast:} & \quad \left[\hat{\mathbf{x}}^{CLSTM}_{i+S_{\text{in}}}, \ldots, \hat{\mathbf{x}}^{CLSTM}_{i+S_{\text{in}}+S_{\text{out}}-1}\right] \xrightarrow{\text{Prediction}} \nonumber \\
& \quad \left[\hat{\mathbf{x}}^{CLSTM}_{i+S_{\text{in}}+S_{\text{out}}}, \ldots, \hat{\mathbf{x}}^{CLSTM}_{i+S_{\text{in}}+2S_{\text{out}}-1}\right] \label{eq:rolling_prediction}
\end{align}

\begin{algorithm}
\caption{Training a physics-constrained ConvLSTM}
\label{algo: Physics-ConvLSTM}
\begin{algorithmic}[1]
\State \textbf{Inputs:} $\mathbf{x}_t$ (state fields), $\mathbf{x}^r_t$ (tessellated observation fields).
\State \textbf{Parameters:} $N_{\text{init}}$ (initial epochs without physics-constrained loss), $N_{\text{epoch}}$ (total epochs), $\eta$ (learning rate), $\theta_{CLSTM}$ (ConvLSTM parameters), $S_{\text{in}}$ (input sequence length), $S_{\text{out}}$ (output sequence length), $\lambda_{\text{energy}}$ (energy conservation loss weight), $T$ (total timesteps), $\mathcal{L}_{total}$ (composite loss function), $\mathcal{E}$ (energy calculation function).
\For{$n = 1$ to $N_{\text{epoch}}$}
    \For{$t = 1$ to $T - S_{\text{in}} - S_{\text{out}} + 1$}
        \State \textbf{Forward Pass via ConvLSTM model:}
        \State $\hat{\mathbf{x}}^{CLSTM}_{t+S_{\text{in}}:t+S_{\text{in}}+S_{\text{out}}-1}$ = \text{ConvLSTM}($\mathbf{x}^r_{t:t+S_{\text{in}}-1}$; $\theta_{CLSTM}$)
        
        \If{$n > N_{\text{init}}$}
            \State \textbf{Compute Physics Constraint Loss:}
            \State $\mathcal{L}_{\text{energy}} = \left| \mathcal{E}\left(\hat{\mathbf{x}}^{CLSTM}_{t+S_{\text{in}}:t+S_{\text{in}}+S_{\text{out}}-1}\right) - \mathcal{E}\left(\mathbf{x}_{t:t+S_{\text{in}}-1}\right) \right|$
        \Else
            \State $\mathcal{L}_{\text{energy}} = 0$
        \EndIf

        \State \textbf{Compute Composite Loss:}
        \State $\mathcal{L}_{\text{total}} = \|\mathbf{x}_{t+S_{\text{in}}:t+S_{\text{in}}+S_{\text{out}}-1} - \hat{\mathbf{x}}^{CLSTM}_{t+S_{\text{in}}:t+S_{\text{in}}+S_{\text{out}}-1}\|^2 + \lambda_{\text{energy}} \cdot \mathcal{L}_{\text{energy}}$
        
        \State \textbf{Update Parameters:}
        \State Backpropagate to compute $\nabla_{\theta_{CLSTM}} \mathcal{L}_{\text{total}}$ and update $\theta_{CLSTM}$ using Adam optimizer with $\eta$
    \EndFor
\EndFor
\end{algorithmic}
\end{algorithm}

As highlighted in Algorithm~\ref{algo: Physics-ConvLSTM}, we integrate physical rules into our training only after initial epochs, a strategy that distinguishes it from typical \ac{LSTM} model training. The \ac{CED-LSTM} model benefits from an early start, as data processing through the \ac{CED} occurs at the outset. This preliminary processing is instrumental in elucidating the relationship between the Voronoi tessellations and the state fields. By handling data at this early stage, the \ac{LSTM} is better poised to enhance both the \ac{MSE} and energy conservation loss assessments right from the beginning, leading to more effective and accurate predictions.

In contrast, the \ac{ConvLSTM} model makes spatio-temporal predictions using Voronoi tessellation inputs directly. Hence, we do not initially integrate the energy loss into the model's training process. Due to \ac{ConvLSTM}'s detailed focus on both spatial and temporal aspects, starting with physical rules could initially hinder the learning of basic data patterns. Therefore, we initially focus on reducing the MSE for the first $N_{\text{init}} = 50$ epochs. After achieving stable and satisfactory reductions in MSE, we then incorporate the physics constraint loss into the training loss function to refine the \ac{ConvLSTM}'s predictions. This gradual, step-by-step approach ensures a balanced learning process. It allows the model to first learn the fundamental structure of the data before aligning with physical laws. This method helps in reducing numerical errors and better reflecting real-world phenomena.

\section{Numerical Example: NOAA Sea Surface Temperature}
\label{sec:experiments-NOAA}

\subsection{Dataset Description and Experimental Setup}

\begin{figure*}[h!]
\centering
\includegraphics[width=0.8\textwidth]{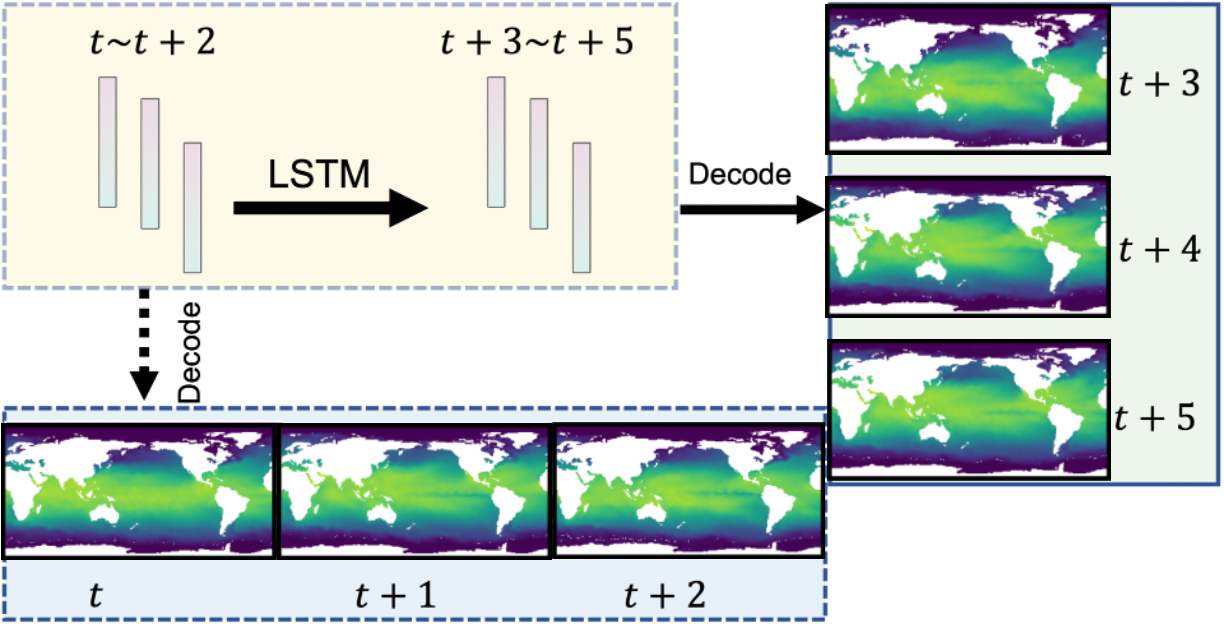}
\caption{Illustration of the \ac{LSTM} process for extracted \ac{NOAA SST} latent space prediction. This figure provides a visual comprehension of the \ac{LSTM} framework's operational mechanism in processing and predicting temporal sequences within the oceanographic context.}
\label{fig:NOAA_LSTM_Process_Prediction}
\end{figure*}

The \ac{NOAA SST} dataset, derived from satellite and ship-based observations, offers critical insights into variations in oceanic temperatures by providing weekly snapshots of sea surface temperatures with a spatial resolution of \(360 \times 180\)~\cite{donlon2007global}. We used data spanning from Year 1981 to 2001 for our training and testing, capturing a broad spectrum of oceanic conditions and temporal changes. Given the weekly nature of \ac{NOAA SST} field changes, using excessively long time steps in our experiments becomes impractical and lacks meaningful interpretation. As illustrated in Figure~\ref{fig:NOAA_LSTM_Process_Prediction}, changes in NOAA SST fields occur rapidly. Therefore, we limited our spatio-temporal prediction experiments to \(S_{in} = S_{out} = 3\), where \(S_{in}\) and \(S_{out}\) represent the input and output temporal sequence lengths, respectively.

As \ac{NOAA SST} dataset represents an irregular dynamical system in the real world without explicit formulas, applying physics constraints to this numerical experiment is not feasible. We will demonstrate the accuracy and robustness of the \ac{DSOVT} framework in this scenario with movable and flexible numbers of sensors for only multi-step prediction.

Sensor placement on water surfaces simulates real-world variability with random positions. This study assesses how sensor density affects sea surface temperature prediction accuracy. The models are trained and tested using Voronoi tessellations with sensor counts ranging from 200 to 340, as detailed in Table~\ref{table:sensor_distribution}. The training configurations use sets of \{200, 240, 280, 320\} sensors, each set tested under three random seeds, totaling 12 cases. For testing, configurations extend to including \{300, 340\} sensors, assessing performance on unseen scenarios. 

\begin{table}[h!]
\centering
\begin{tabular}{c|c|c} \toprule
    \textbf{Dataset} & \textbf{Number of Sensors (Snapshots)} & \textbf{Percentage of Grid Points} \\ \midrule
    \multirow{4}{*}{Training} & 200 sensors (1040 $\times$ 3) & 0.31\% \\
                              & 240 sensors (1040 $\times$ 3) & 0.37\% \\
                              & 280 sensors (1040 $\times$ 3) & 0.43\% \\
                              & 320 sensors (1040 $\times$ 3) & 0.49\% \\ \midrule
    \multirow{6}{*}{Testing}  & 200 sensors (1040 $\times$ 1) & 0.31\% \\
                              & 240 sensors (1040 $\times$ 1) & 0.37\% \\
                              & 280 sensors (1040 $\times$ 1) & 0.43\% \\
                              & 300 sensors (1040 $\times$ 1) & 0.46\% \\
                              & 320 sensors (1040 $\times$ 1) & 0.49\% \\
                              & 340 sensors (1040 $\times$ 1) & 0.52\% \\
    \bottomrule
\end{tabular}
\caption{Summary of sensor configurations and seed settings in training and testing datasets. The number of sensors and the corresponding percentage of grid points covered are shown. Training data at each sensor density were derived from the 1040 snapshots in the NOAA SST dataset using three distinct random seeds (300, 100, and 10). Testing data were uniquely generated using a different random seed (900) for each sensor density, with each sensor density consisting of 1040 snapshots.}
\label{table:sensor_distribution}
\end{table}

Our experiment employed \ac{CED-LSTM} and \ac{ConvLSTM} models trained on the \ac{NOAA SST} dataset to predict global temperature fields based on sparse and time-varying sensor data. 
In the 2D-Kriging experiment, we applied the OrdinaryKriging function from PyKrige package with a Spherical variogram model and set the number of averaging bins for the semivariogram to 4 for spatial interpolation, followed by a linear regression for temporal prediction. The 3D-Kriging experiment used the OrdinaryKriging3D function with the same variogram model, but increased the number of averaging bins to 5. In this setup, we provided spatio-temporal information of sparse fields for the initial three steps, utilizing 3D-Kriging to predict the spatio-temporal interpolation results for the subsequent three steps.

For the \ac{NOAA SST} numerical experiments, the \ac{ReLU} function is selected as the activation function for \ac{CED-LSTM} and \ac{ConvLSTM} due to their compatibility with the field range.

\subsection{DSOVT (CED-LSTM)}

\subsubsection{Validation of CED: From Voronoi Tessellations to State Fields}

The \ac{CED} transitions from sparse inputs to state field outputs, demonstrating its ability to comprehend complex spatial relationships and infer missing information. This capability is particularly valuable in oceanographic data analysis, where conventional methods may struggle due to the sparsity of data and its time-varying nature~\cite{george2021deep}. The effectiveness of the \ac{CED} model's latent space representation in capturing essential information is validated through our quantitative evaluations, as reflected in the \ac{SSIM}, \ac{PSNR} and \ac{R-RMSE}. For \ac{NOAA SST} dataset,  we choose $Z = 512$ as the dimension of the latent space in our \ac{CED} model.

\begin{figure*}[htbp]
\centering
\includegraphics[width=\textwidth]{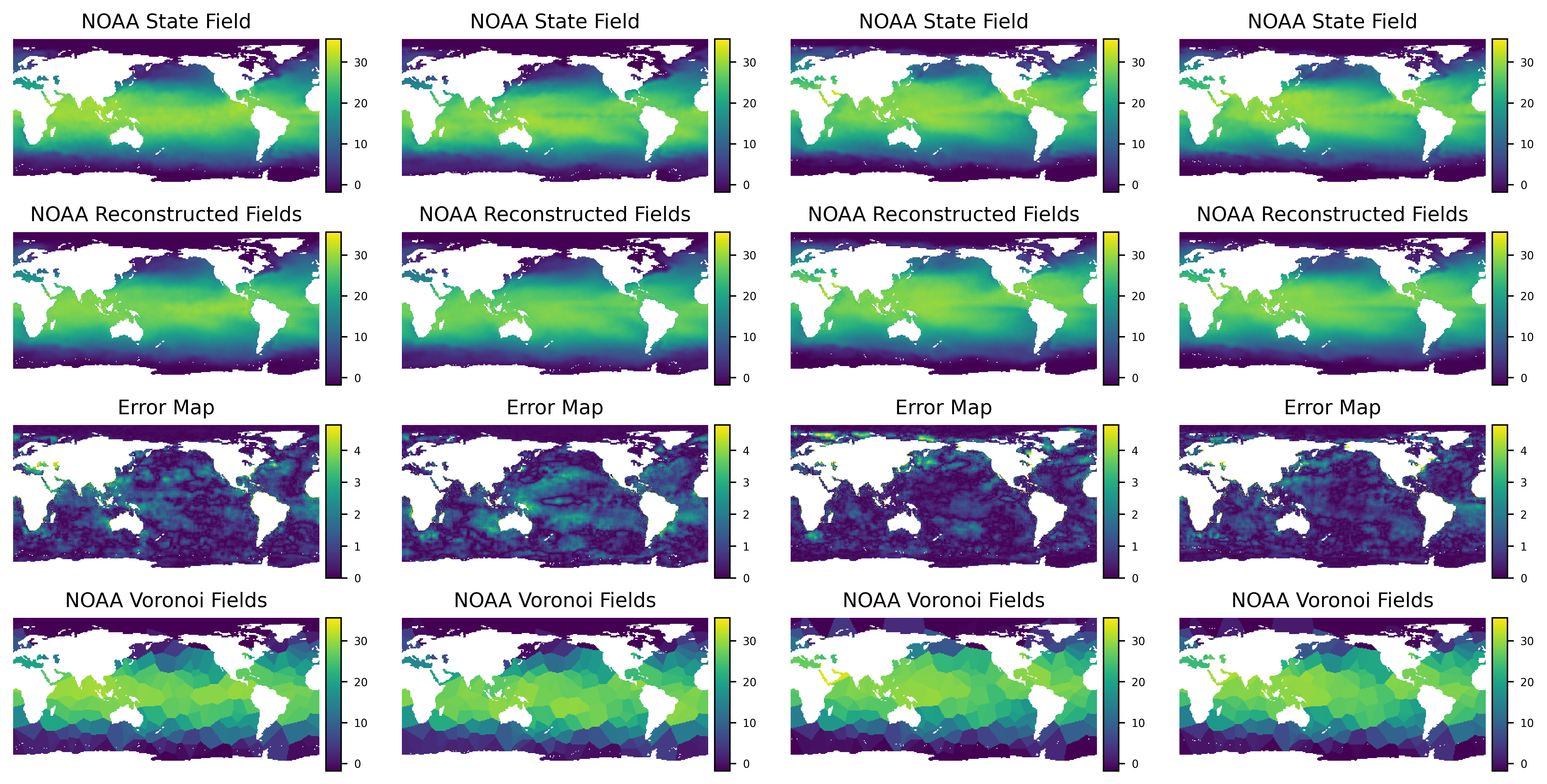}
\caption{Reconstructed Fields from Voronoi tessellation, including the actual fluid dynamics visualization. The figure is organized into four rows depicting: (1) NOAA state fields, (2) CED reconstructed fields from Voronoi tessellations, (3) Error maps showing discrepancies between state and predicted fields, and (4) Voronoi tessellations derived from 200 time-varying sensor data. Each column represents a different step.}
\label{fig:NOAA_CED_Prediction}
\end{figure*}

Firstly, we present the efficacy of our CED in handling Voronoi tessellation fields. Figure~\ref{fig:NOAA_CED_Prediction} underscores the efficacy of our CED in transforming Voronoi tessellations into state fields with 200 time-varying sensors. As shown in Figure~\ref{fig:NOAA_CED_Prediction}, sensors are randomly deployed across the ocean surface, and based on the data collected from these sensors, Voronoi tessellations are generated. These tessellations serve as inputs to the CED, which is used to reconstruct the state fields. By comparing the state fields reconstructed by the CED with the actual state fields, we ensure that the latent representation contains information that is both accurate and effective.
The \ac{CED} achieves an \ac{SSIM} of 0.93, a \ac{PSNR} of 32.18 dB and R-RMSE of 0.09 on our test dataset, which are indicators of its proficiency in reconstructing \ac{NOAA SST} fields. These metrics confirm the CED's capability to generate fields that are both structurally and visually similar to the state fields.

\subsubsection{Accuracy and Efficiency}
\ac{LSTM} models demonstrate remarkable proficiency in bridging gaps in sequential data, which makes them particularly adept for applications in environmental data analysis, such as \ac{NOAA SST} prediction. This section shows the \ac{LSTM}'s application in modeling and predicting the latent space dynamics inherent in \ac{NOAA SST} data.

\begin{figure*}[htbp]
\centering
\includegraphics[width=\textwidth]{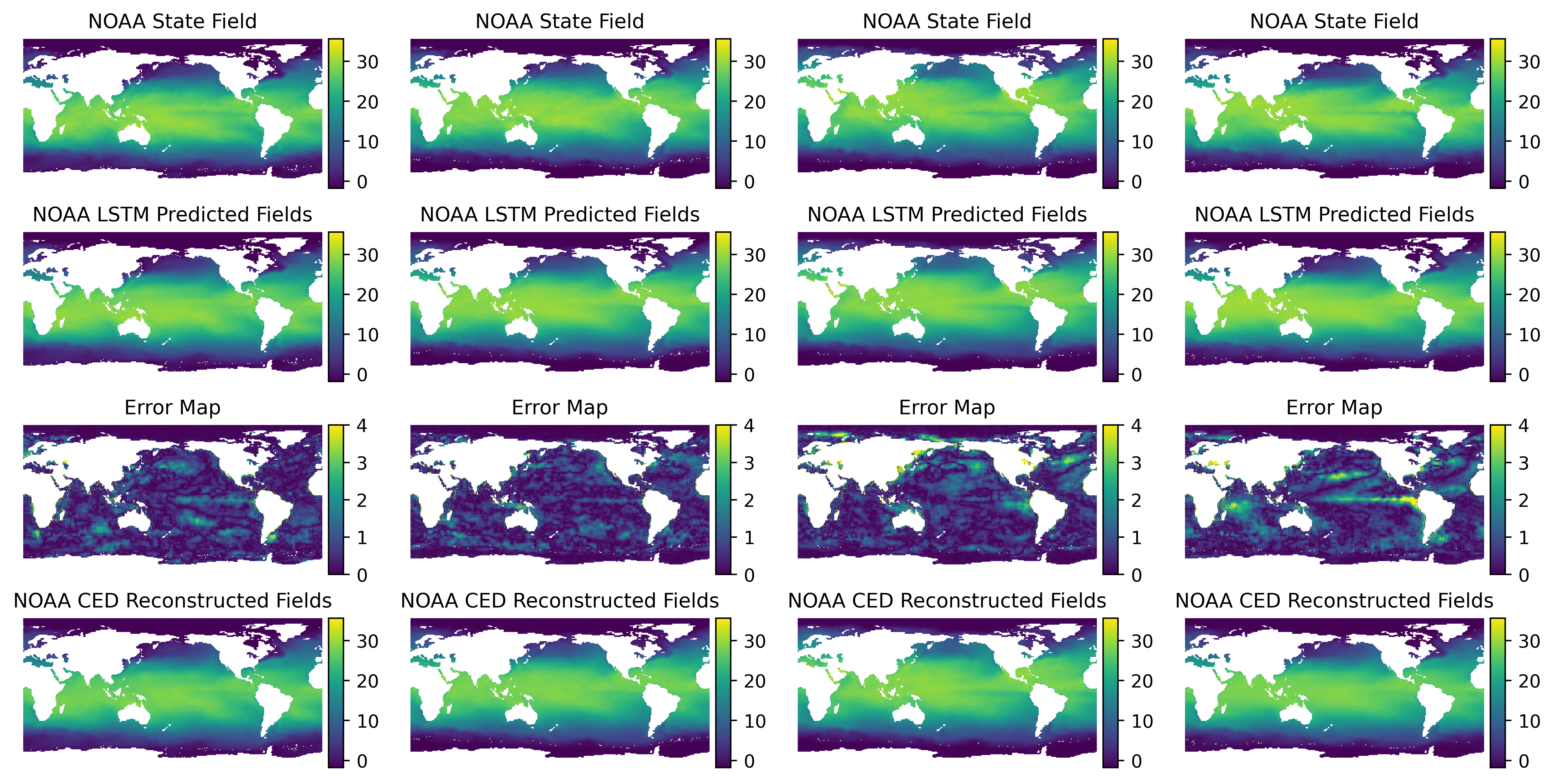}
\caption{Comparison of \ac{NOAA SST} state fields with \ac{LSTM} and CED-based predictions across different steps. Each column represents a progression of steps: Step 10, Step 50, Step 90, and Step 130. The arrangement within each column follows this order: the first row displays the NOAA SST state field, the second row shows LSTM predictions, the third row depicts the error maps comparing the LSTM predictions to the state field, and the fourth row presents the CED-based reconstructions.}
\label{fig:NOAA_LSTM_Prediction}
\end{figure*}

When evaluating the performance of the models in terms of accuracy and efficiency, the \ac{LSTM} exhibits superior capabilities. As shown in Table~\ref{tab:comparison}, the integrated \ac{CED-LSTM} framework achieves an \ac{SSIM} of 0.84, a \ac{PSNR} of 36.34 dB and a R-RMSE of 0.08, while all maintaining an inference time of 2.47 seconds. In contrast, the Kriging method, which records the highest \ac{SSIM} of 0.69, \ac{PSNR} of 25.08 dB and R-RMSE of 0.31, is markedly surpassed by our model, which shows an improvement of 20.64\% in \ac{SSIM}, 44.89\% in \ac{PSNR} and 74.19\% in \ac{R-RMSE}. In addition, detailed in Figure~\ref{fig:NOAA_LSTM_Prediction} is a comparative analysis showcasing the \ac{NOAA SST} state fields against \ac{CED-LSTM} predictions and CED reconstructions, alongside \ac{CED-LSTM} respective error mapping. These comparative illustrations and metrics collectively underscore the \ac{LSTM} model's capability in predicted \ac{NOAA SST} data with high fidelity and computational efficiency.

\subsection{DSOVT (ConvLSTM)}
\subsubsection{Accuracy and Efficiency}

\begin{figure*}[h!]
\centering
\includegraphics[width=1.0\textwidth]{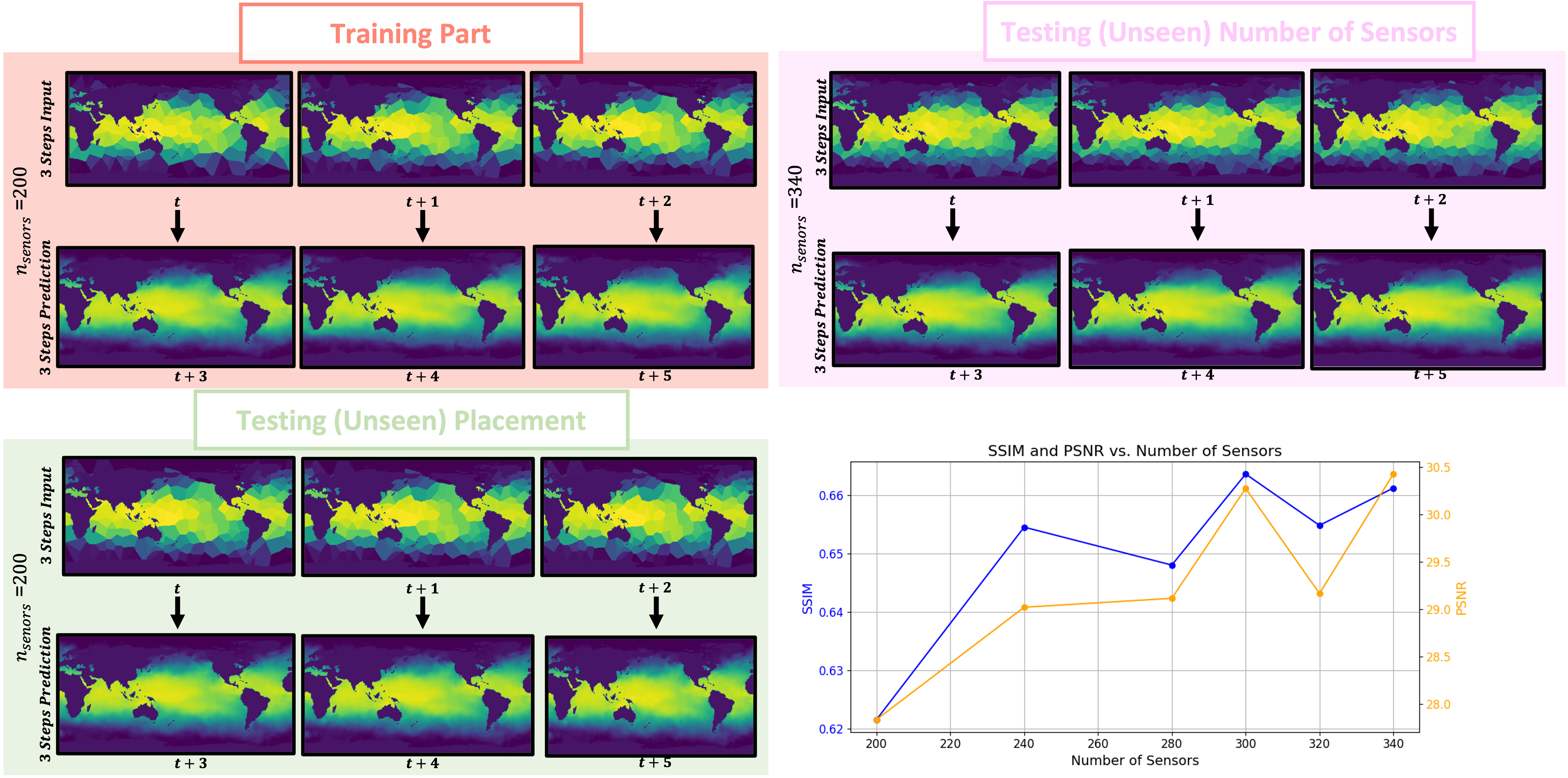}
\caption{Voronoi-based spatial data prediction of NOAA sea surface temperature using \ac{ConvLSTM}. This figure illustrates the training process with sensor counts set at 200, matching the number of sensors in the training data, and extends to cases with sensor counts of {200, 340} that are not present in the training dataset. The line graph depicts, within the test dataset, the performance of the \ac{ConvLSTM} model in terms of SSIM and PSNR across different sensor counts.}
\label{fig:convlstm-NOAA-compare}
\end{figure*}

The \ac{ConvLSTM} model exhibits good capability in directly predicting \ac{NOAA SST} fields from Voronoi tessellation inputs through an integrated end-to-end pipeline. This streamlined approach enables the efficient utilization of sparse, dynamically distributed oceanic sensor data, setting it apart from the \ac{CED-LSTM}, which necessitates the training of two distinct models for handling Voronoi tessellation inputs and generating accurate field predictions. In our experiments, we normalized the \ac{NOAA SST} data to ensure consistency in scale and facilitate model convergence.

\begin{figure*}[htbp]
\centering
\includegraphics[width=\textwidth]{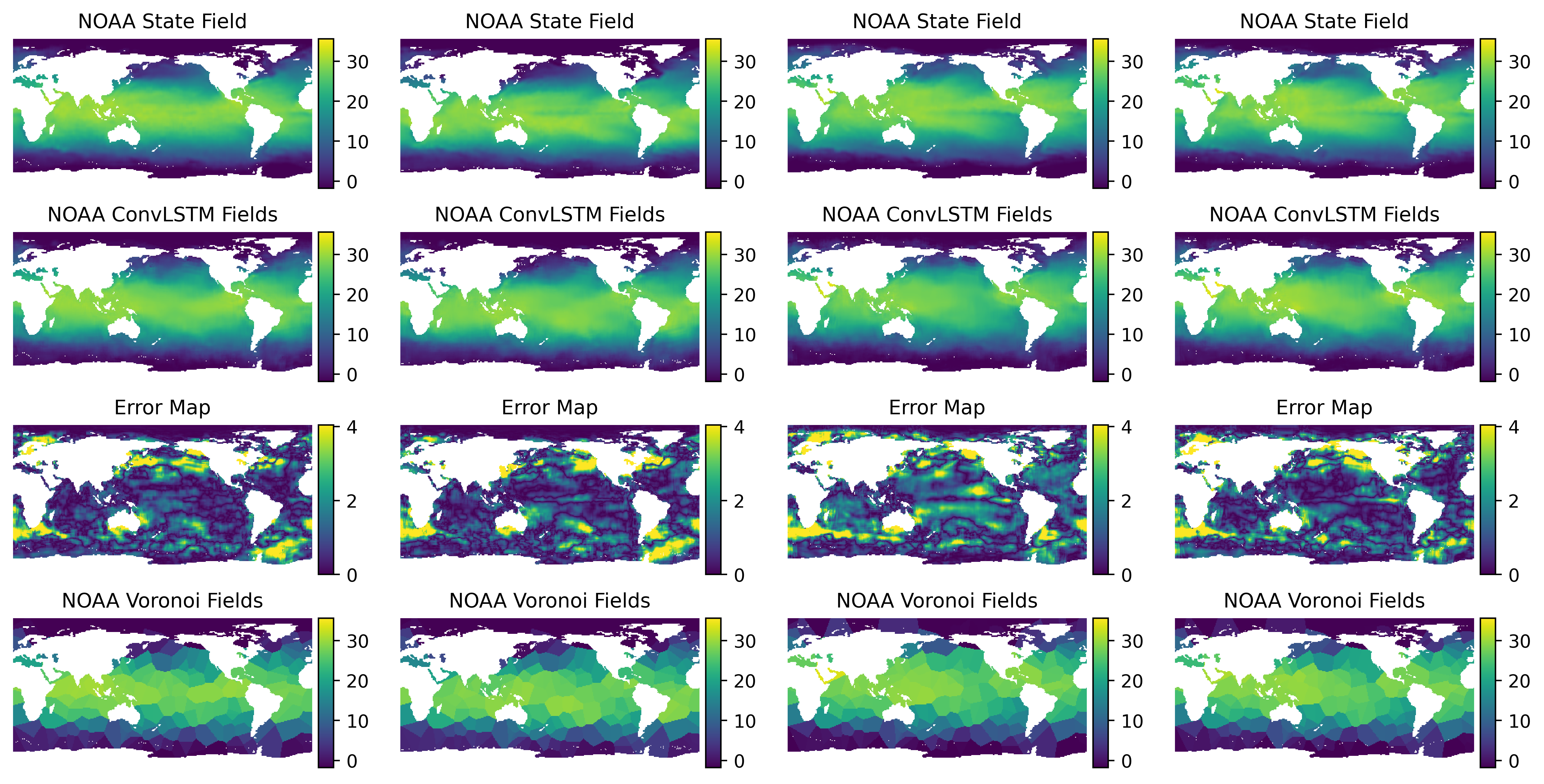}
\caption{Comparison of \ac{NOAA SST} state fields with ConvLSTM across different steps. Each column represents a progression of steps: Step 80, Step 120, Step 200, and Step 400. The arrangement within each column follows this order: the first row displays the NOAA SST state field, the second row shows ConvLSTM predictions, the third row depicts the error maps comparing the ConvLSTM predictions to the state field, and the fourth row presents the corresponding Voronoi tessellations.}
\label{fig:NOAA_convlstm_Prediction}
\end{figure*}

In our multi-step prediction analysis, as shown in Table~\ref{tab:comparison}, the \ac{ConvLSTM} model achieves \ac{SSIM}, \ac{PSNR} and \ac{R-RMSE} scores of 0.75, 29.59 dB and 0.13, respectively, in our test datasets. The performance metrics for different numbers of sensors in test datasets are illustrated in the line graph in Figure~\ref{fig:convlstm-NOAA-compare}.
These metrics underscore the \ac{ConvLSTM}'s capability to accurately predict \ac{NOAA SST} fields, demonstrating a substantial improvement over \ac{2D-Kriging}. Specifically, the increases are 22.95\% in \ac{SSIM} and 17.98\% in PSNR. As shown in Figures~\ref{fig:NOAA_convlstm_Prediction} and~\ref{fig:representative_NOAA_Prediction}, \ac{ConvLSTM} not only effectively predicts NOAA SST fields but also outperforms traditional Kriging methods, particularly in terms of capturing high-resolution features and minimizing errors.

It should be noted that the inference time for \ac{ConvLSTM} is 33.14 seconds, which is longer compared to \ac{CED-LSTM} and \ac{2D-Kriging}. This extended duration is attributed to \ac{ConvLSTM}'s model parameter size. Meanwhile, \ac{ConvLSTM}'s accuracy of predictions is lower than our \ac{CED-LSTM} as shown in Table~\ref{tab:comparison}. This may be because \ac{CED-LSTM} is able to perform spatial reconstruction on Voronoi tessellations before making temporal predictions in the latent space. In contrast, \ac{ConvLSTM} captures spatial dependencies and handles temporal changes simultaneously but benefits from an end-to-end training structure, which is more practical in real-world dynamical systems.

\subsection{Comparative Analysis with Alternative Approaches}

\begin{table}[h!]
    \centering
    \begin{tabular}{@{}lcccc@{}}
        \toprule
        Model & SSIM & PSNR (dB) & R-RMSE & Inference Time (s)\\ 
        \midrule
        2D-Kriging & 0.61 & 25.08 & 0.31&4.76 \\
        3D-Kriging & 0.69 & 20.93 &0.35 &11154.35 \\ 
        CED-LSTM & 0.84 & 36.34 & 0.08&2.47 \\
        ConvLSTM & 0.75 & 29.59 &0.13 & 33.14 \\
        \bottomrule
    \end{tabular}
    \caption{Comparative assessment of different models' performance in multi-step predictions on the NOAA SST testing datasets, evaluated using SSIM, PSNR, R-RMSE and inference time in seconds.}
    \label{tab:comparison}
\end{table}

\begin{figure*}[h!]
    \centering
    \includegraphics[width=1.0\textwidth]{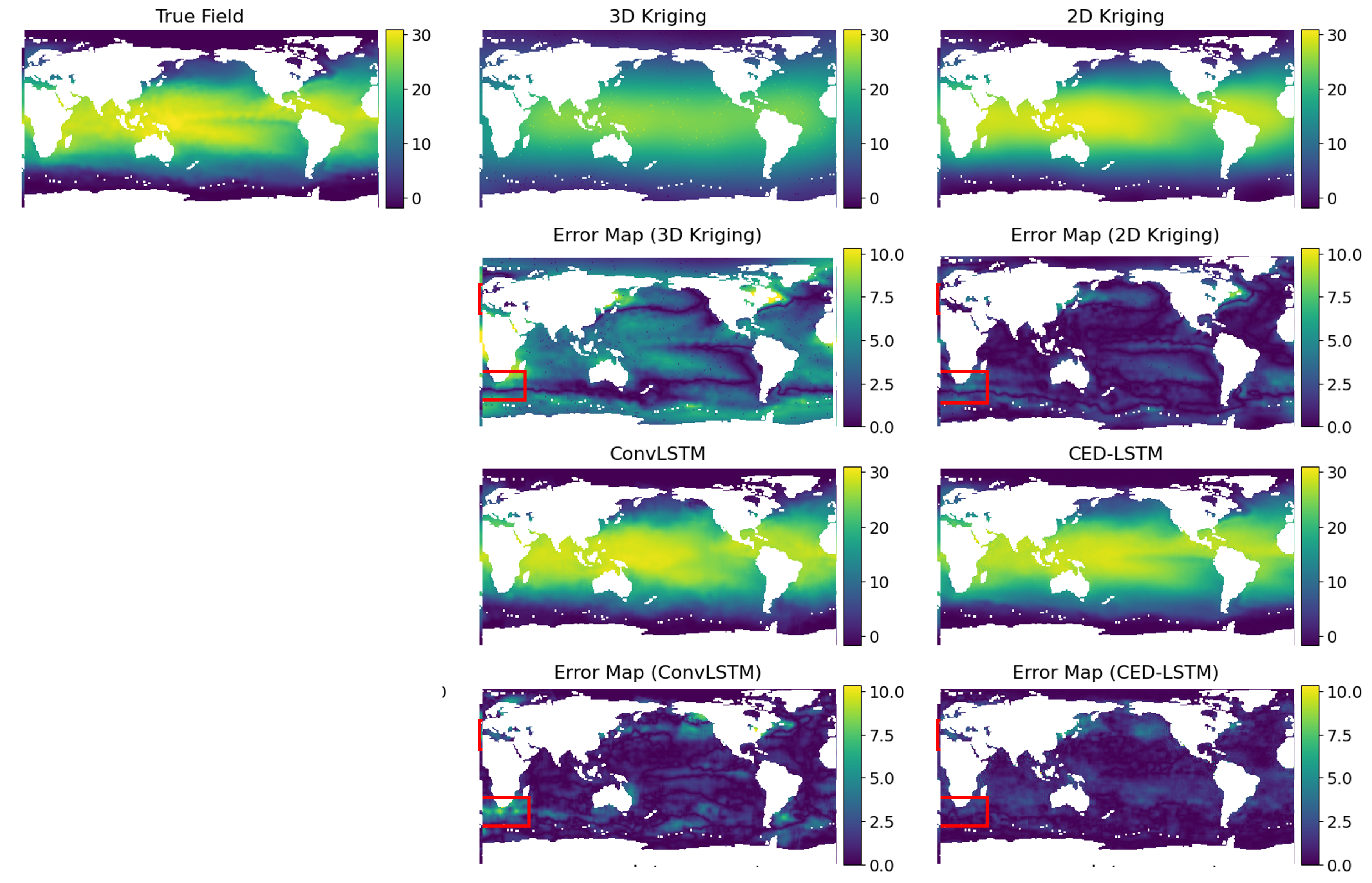}
    \caption{Detailed comparison of multi-step prediction outputs from different models. The first row showcases the state field alongside predictions from 3D Kriging and 2D Kriging, followed by their respective error maps. The third row presents ConvLSTM and CED-LSTM predictions, with the final row depicting error maps for ConvLSTM, and CED-LSTM models.}
    \label{fig:representative_NOAA_Prediction}
\end{figure*}

Utilizing the DSOVT framework within the NOAA SST dataset, the integration of CED-LSTM and \ac{ConvLSTM} models has outperformed conventional 2D-Kriging and 3D-Kriging techniques, focusing on key metrics: PSNR, SSIM, R-RMSE, and inference time. From Figure~\ref{fig:representative_NOAA_Prediction}, we can see that compared to the error maps of Kriging methods, those of ConvLSTM and CED-LSTM show significant improvement. Specifically, CED-LSTM performs better with an SSIM of 0.84 and a PSNR of 36.34 dB, as highlighted in Table~\ref{tab:comparison}, illustrating its superior capability in preserving structural integrity and fine details essential for precise spatio-temporal evaluations. Meanwhile, its R-RMSE is 0.08, indicating a low error rate relative to the range of the state fields.
Moreover, its quick inference time of 2.47 seconds meets the needs of real-time analysis, highlighting its suitability for applications that require fast and accurate spatio-temporal predictions. In contrast, ConvLSTM model showcases its robust potential, particularly excelling with its end-to-end architecture, which simplifies the training process, unlike the CED-LSTM model, which requires separate training phases for each component. 

Conversely, \ac{2D-Kriging}, while faster, fails to accurately model complex spatio-temporal interactions, as shown by its lower \ac{SSIM}, \ac{PSNR}, and higher R-RMSE scores. This limits its effectiveness in detailed spatial analysis. \ac{3D-Kriging}, despite its higher \ac{SSIM}, struggles with efficient use of spatial features (PSNR = 20.93 dB) and requires a lot of computing time (11154.35 s), which limits its practical use. These assessments highlight the crucial trade-off between computational speed and accuracy in spatio-temporal prediction methods.

\section{Numerical Example: Shallow Water Systems}
\label{sec:experiments-SW}
\subsection{Dataset Description and Experimental Setup}
Shallow water systems provide insightful models to understand non-linear wave phenomena~\cite{osborne1998solitons}. Our analysis begins with a cylindrical disturbance introduced into water at time $t=0$, deliberately excluding the Coriolis force to prioritize the study of horizontal over vertical scales. This approach directs us towards the application of the Saint-Venant equations, fundamental principles in fluid mechanics formulated by Saint-Venant in the nineteenth century~\cite{Venant1871}. These equations elegantly describe the relationship between horizontal fluid velocity and fluid height as follows:

\begin{align}
    & \frac{\partial h}{\partial t} + \frac{\partial(hu)}{\partial x} + \frac{\partial(hv)}{\partial y} = 0, \\
    & \frac{\partial(hu)}{\partial t} + \frac{\partial\left(hu^2 + \frac{1}{2}gh^2\right)}{\partial x} + \frac{\partial(huv)}{\partial y} = 0, \\
    & \frac{\partial(hv)}{\partial t} + \frac{\partial(huv)}{\partial x} + \frac{\partial\left(hv^2 + \frac{1}{2}gh^2\right)}{\partial y} = 0.
\end{align}

In these equations, \(u\) and \(v\) represent the velocity components of the fluid in the horizontal and vertical directions, respectively, both measured in meters per second \((\mathrm{m/s})\). The variable \(h\) denotes the total water depth in meters \((\mathrm{m})\), including the undisturbed water depth. The gravitational acceleration constant \(g\) is normalized to 1 for the purposes of these simulations, simplifying the dynamics under controlled conditions.

\begin{figure*}[h!]
    \centering
    \includegraphics[width=0.9\textwidth]{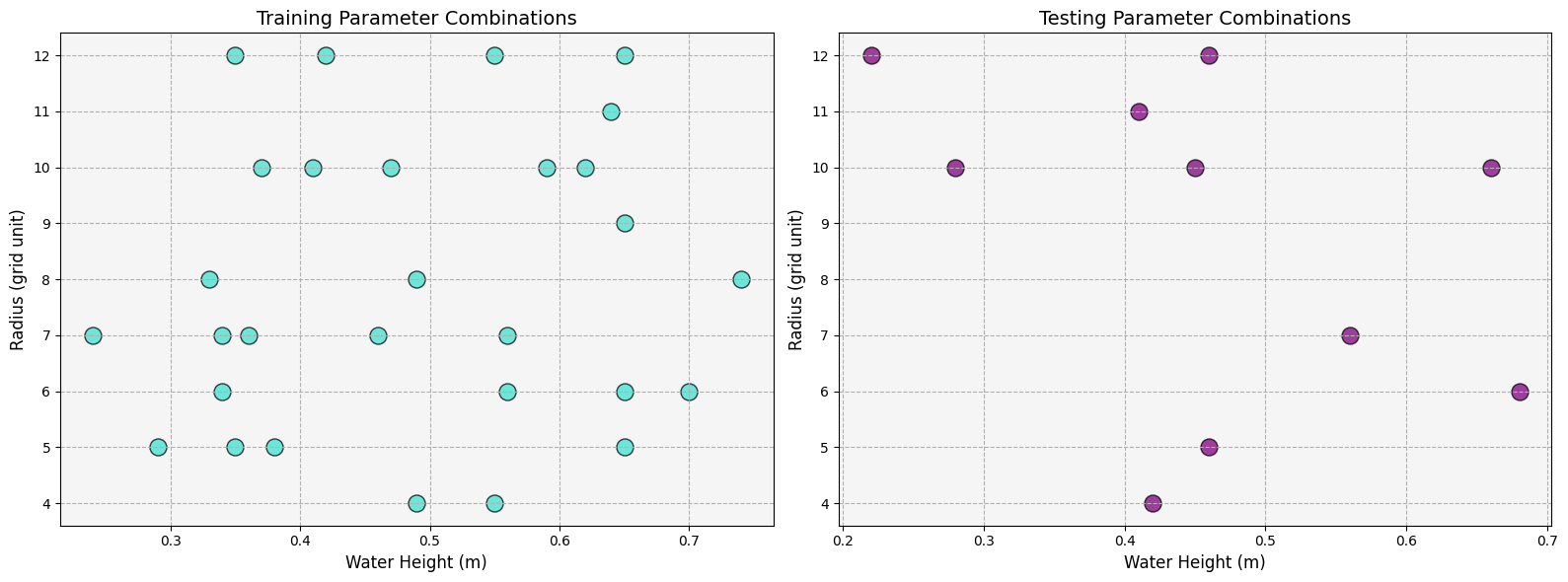}
    \caption{Parameter selection for training and testing datasets in the study of shallow water systems. The left side of the image illustrates the parameters for the training dataset, while the right side corresponds to the test dataset.}
    \label{fig:para_SW}
\end{figure*}

Our computational domain consists of a \(64 \times 64\) grid, with each cell containing three channels for the velocity components \(u, v\), and the water height \(h\). The initial conditions introduce a cylindrical disturbance in the water height, with variations ranging from \([0.2, 0.8] \mathrm{m}\) and a radius \(r\) between \([4, 12]\) grid units. This configuration enables a comprehensive study of wave dynamics and fluid behaviors. The undisturbed water depth throughout the domain is set to \(1 \mathrm{m}\), ensuring consistent baseline conditions for all simulations. Random sampling was employed to select the heights and radius within the specified ranges. As illustrated in Figure~\ref{fig:para_SW}, our dataset comprises 30 simulations for training and 10 for testing. This allocation allows for robust model training and effective validation on unseen data.

For these simulations, we employ a finite difference of the first order method to solve the Saint-Venant equations numerically. The simulations are designed to run for a total of 3500 steps, with an initial phase dedicated to establishing equilibrium that lasts for the first 500 steps. During the simulation, data is collected at regular intervals. Specifically, a snapshot of the system's state is taken every 10 steps, which results in a total of 300 distinct spatio-temporal timesteps being recorded for each simulation. This rigorous sampling regime ensures detailed tracking of the dynamic response to the cylindrical disturbance.

We set \(S_{\text{in}} = S_{\text{out}} = 5\) for all analyses to ensure consistency across our experimental conditions. To simulate sparse and time-varying sensors in shallow water fields, we uniformly placed 100 sensors across the field and then randomly moved them up to two units in all directions (up, down, left, and right).

In the 2D-Kriging and 3D-Kriging experiments, we use a Spherical variogram model and set the number of averaging bins for the semivariogram to 20. This approach helps maintain methodological consistency across the two types of Kriging analyses.

The custom activation function for shallow water systems aligns with the dynamic range and characteristics of the CED data for field reconstruction and ConvLSTM for spatio-temporal prediction. The system includes three channels, denoted by \(N_c = 3\). This activation function operates selectively based on the physical properties associated with each channel index \(n_c\). Specifically, for \(n_c = 1\) and \(n_c = 2\), the \ac{tanh} operation is applied to normalize the velocity components \(u\) and \(v\), ensuring the values are confined within the range \([-1, 1]\). For \(n_c = 3\), the activation function employs \ac{ReLU} to ensure that the water height \(h\) maintains non-negative values, aligning with physical reality where height cannot be negative.

\subsection{DSOVT (CED-LSTM)}

\subsubsection{Validation of CED: From Voronoi Tessellations to State Fields}
\begin{figure*}[h!]
\centering
\includegraphics[width=0.9\textwidth]{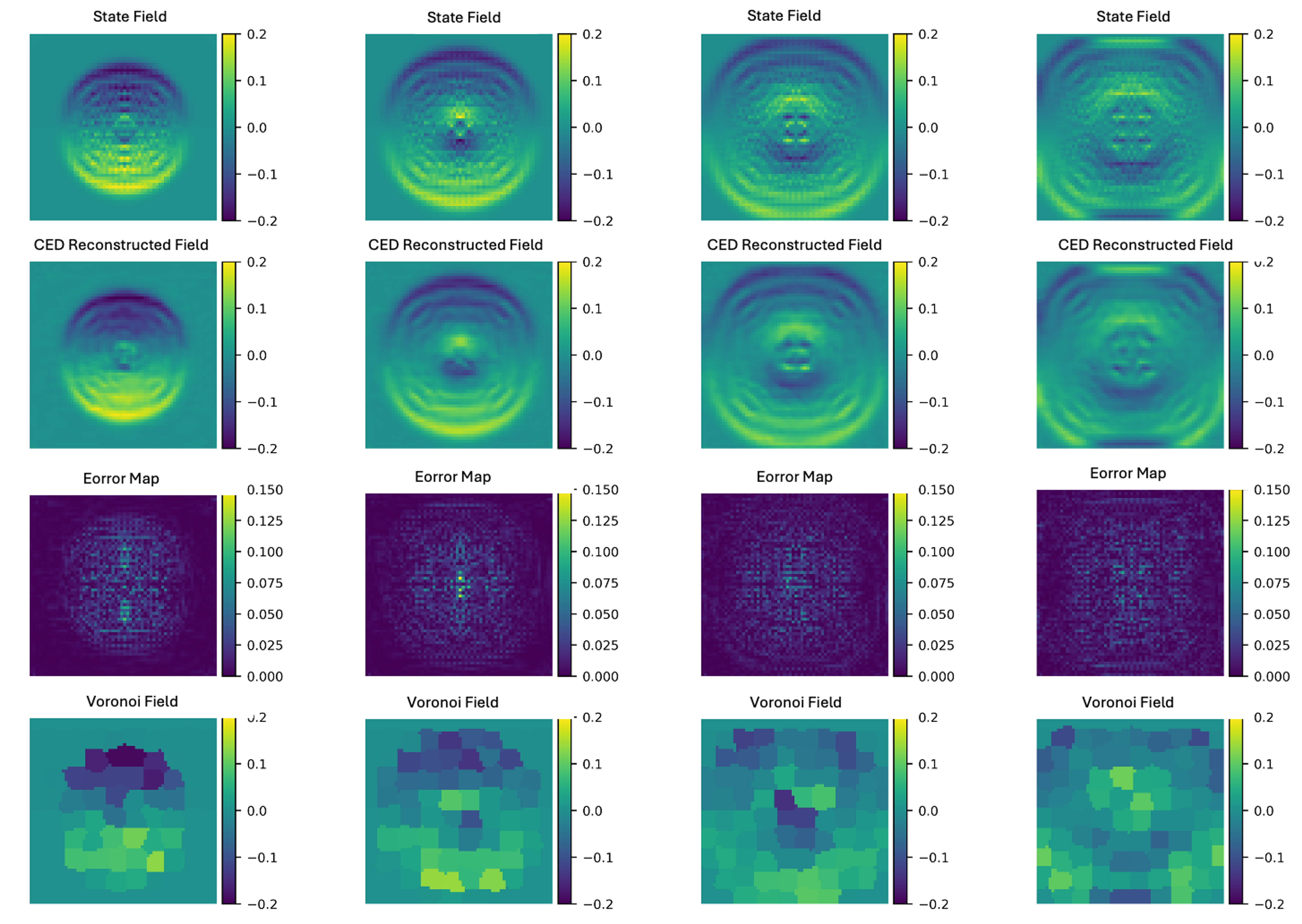}
\caption{Illustration of the CED model's reconstructed performance across different stages of fluid dynamics prediction. The visualization is organized into four rows: (1) Initial state fields of the shallow water system; (2) CED reconstructed fields from Voronoi Tessellation inputs; (3) Error maps comparing the state and reconstructed fields; (4) Corresponding Voronoi tessellation for each case. Each column presents a representative scenario within a shallow water system, showcasing the model's precision in capturing complex fluid dynamics.}
\label{fig:CED_Prediction}
\end{figure*}

First, we demonstrate the robustness and efficacy of the \ac{CED} model on shallow water systems. The dimension of the latent space, $Z$, is set to 128. As illustrated in Figure~\ref{fig:CED_Prediction}, the reconstructed fields accurately capture the feature representations. From the Voronoi tessellation depicted in the fourth row, we observe the \ac{CED} model's capability to reconstruct true state fields from sparsely interpolated fields. It achieves an \ac{SSIM} of 0.95, a \ac{PSNR} of 42.40 dB, and an \ac{R-RMSE} of 0.05, indicating a high degree of structural similarity between the output and the ground truth. These metrics not only highlight the precision of the model but also its effectiveness in addressing data sparsity. 

\subsubsection{Accuracy and Efficiency}

\begin{figure*}[h!]
\centering
\includegraphics[width=0.9\textwidth]{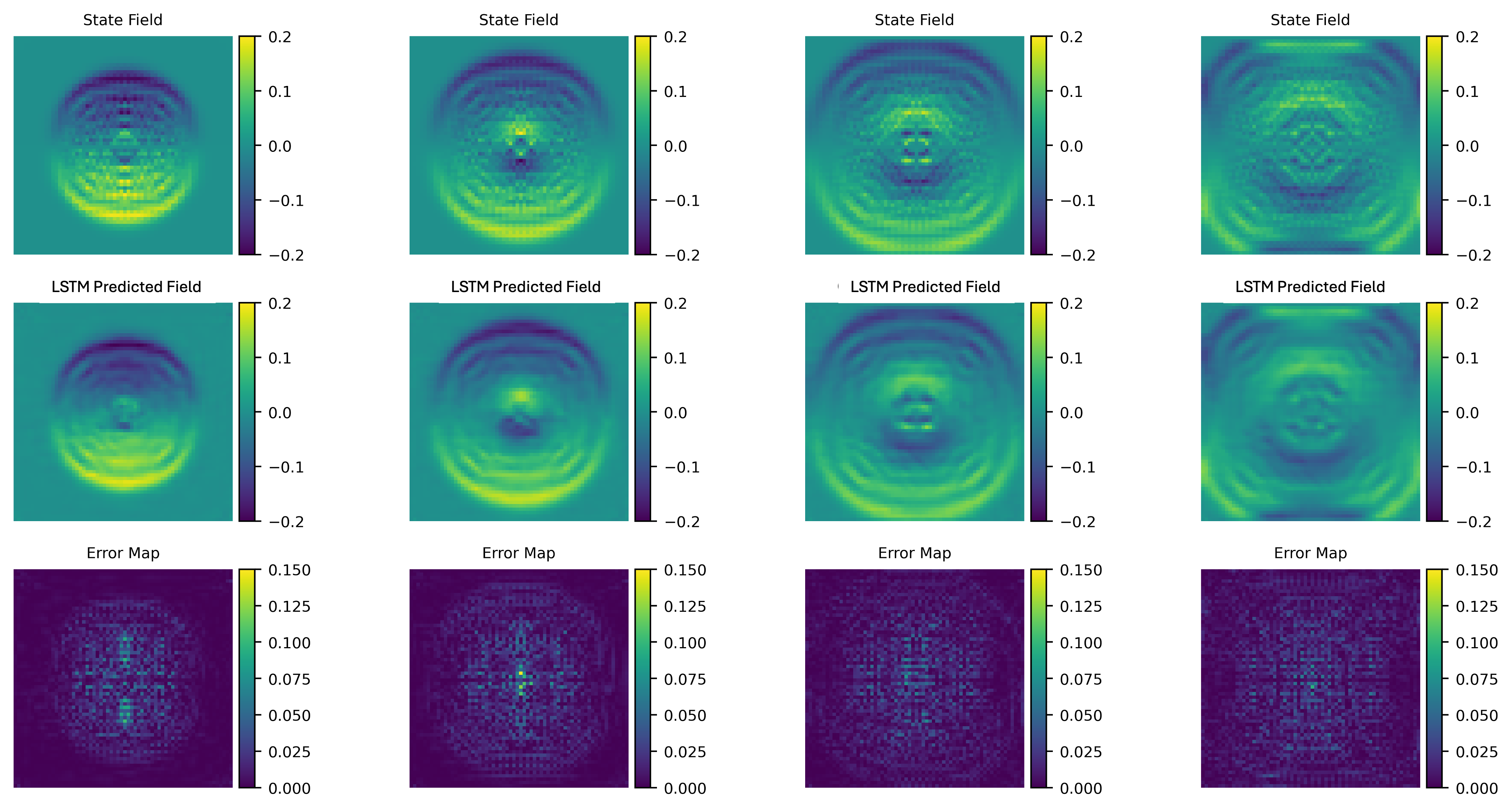}
\caption{Illustration of the LSTM model's predictive performance across different stages of fluid dynamics prediction. The depiction follows a progression from Voronoi tessellation to \ac{LSTM} prediction, culminating in the state fluid dynamics fields. The visualization comprises four rows: (1) Initial state fields of the shallow water system; (2) Predicted fields via LSTM with CED latent representation inputs; (3) Error maps comparing the state and predicted fields. Each column represents a representative shallow water system scenario, showcasing the model's accuracy in capturing detailed fluid dynamics.}
\label{fig:LSTM_Prediction_CED}
\end{figure*}
\begin{figure*}[!h]
\centering
\includegraphics[width=1\textwidth]{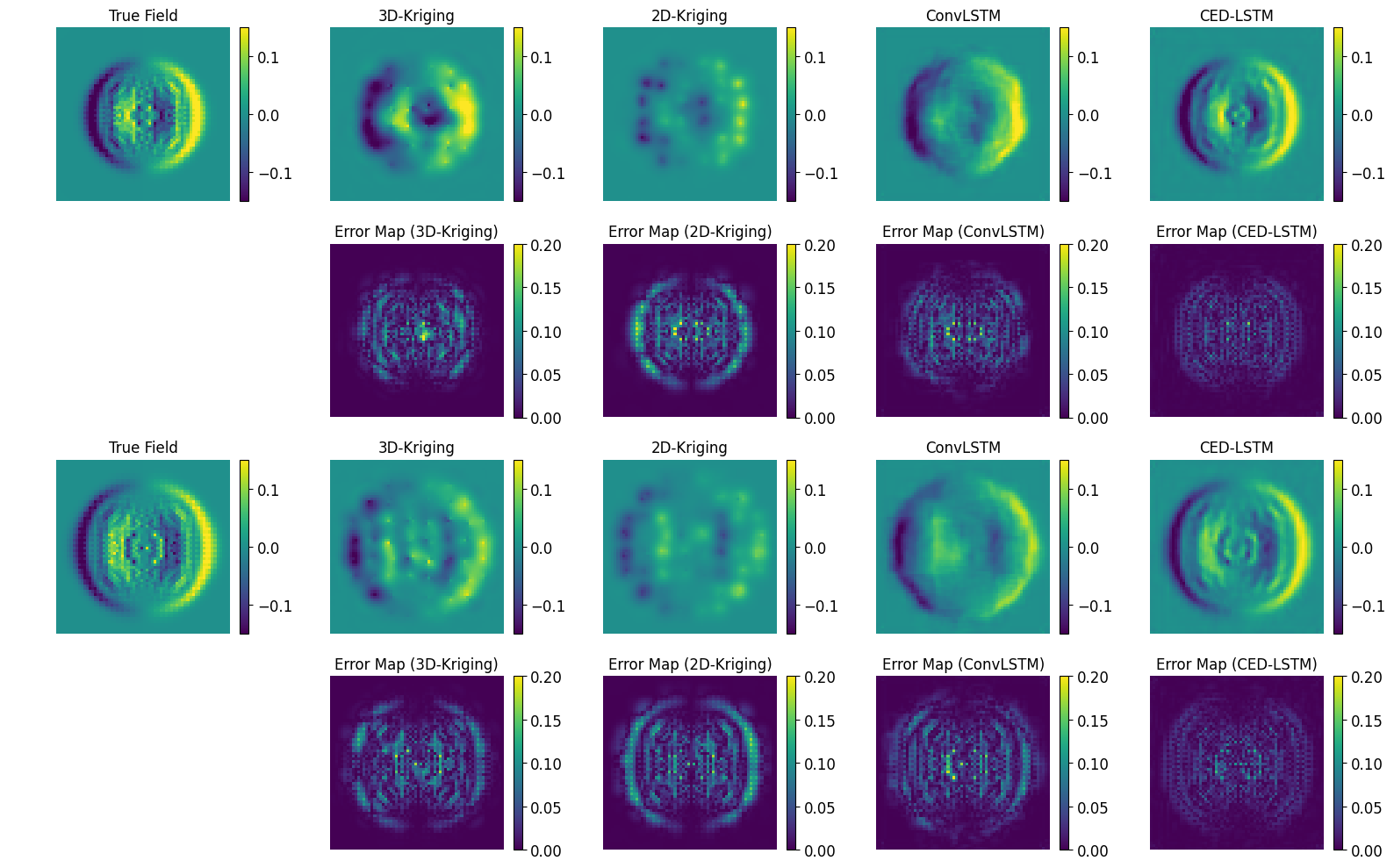}
\caption{Comparison of ground truth fields with predictions from various models, including 3D-Kriging, 2D-Kriging, ConvLSTM, and CED-LSTM. Each row presents the true field, followed by predictions from each model and their respective error maps. The error maps highlight the absolute differences, underlining areas where each model deviates from the observed data. Shown here are two cases: the 105th and 145th steps from a testing simulation, generated with a height of 0.39 m and a radius of 11 grid units.}
\label{fig:all-compare}
\end{figure*}

As shown in Table~\ref{tab:SW-model_performance}, our LSTM-predicted decoded fields achieve an SSIM of 0.96, representing an improvement of approximately 31\% over 2D-Kriging. Furthermore, the PSNR reaches 42.91 dB, indicating a 72.93\% improvement compared to 2D-Kriging. Additionally, the R-RMSE reaches 0.05, which is less than 10\%, demonstrating excellent prediction accuracy. As illustrated in Figure~\ref{fig:LSTM_Prediction_CED}, the predicted fields closely approximate the true fields in terms of feature representations and similarity. As shown in Figure~\ref{fig:all-compare}, it is evident that Kriging underperforms in its predictions of sea surface temperatures in \ac{NOAA SST} experiments. This is likely due to the complex, chaotic nature of features in shallow water systems versus the smoother \ac{NOAA SST} data~\cite{dijkstra2005low}. Kriging assumes stationary spatial variability, which often fails in dynamic environments. Additionally, Kriging's interpolation accuracy depends heavily on sample density~\cite{sirayanone1988comparative}; a sample density of only 2.44\% is insufficient for effective semivariogram estimation.

\subsubsection{Effect of energy conservation constraints on CED-LSTM}
Our investigation extends to exploring the influence of physics constraints on the rolling forecasts made by \ac{CED-LSTM}. In our analysis, we utilized the first 20 training simulations for training, without making any changes to the test dataset. Additionally, we configured the latent representation $Z$ to 64 dimensions to demonstrate the assistance provided by energy conservation constraints in rolling forecasts under scenarios of insufficient sample size in real physical systems. Here we choose $\lambda_{\text {energy}} = 5e-10$ as our weight coefficient of the energy constraint. The process of rolling forecasts was initiated at step 75, marking the beginning of the forecast period within each of the 10 distinct test simulations. We choose to start at step 75 to ensure that the predictions are based on dynamic processes that have sufficiently evolved from the initial conditions.

\begin{figure*}[ht!]
    \centering
    \includegraphics[width=0.7\linewidth]{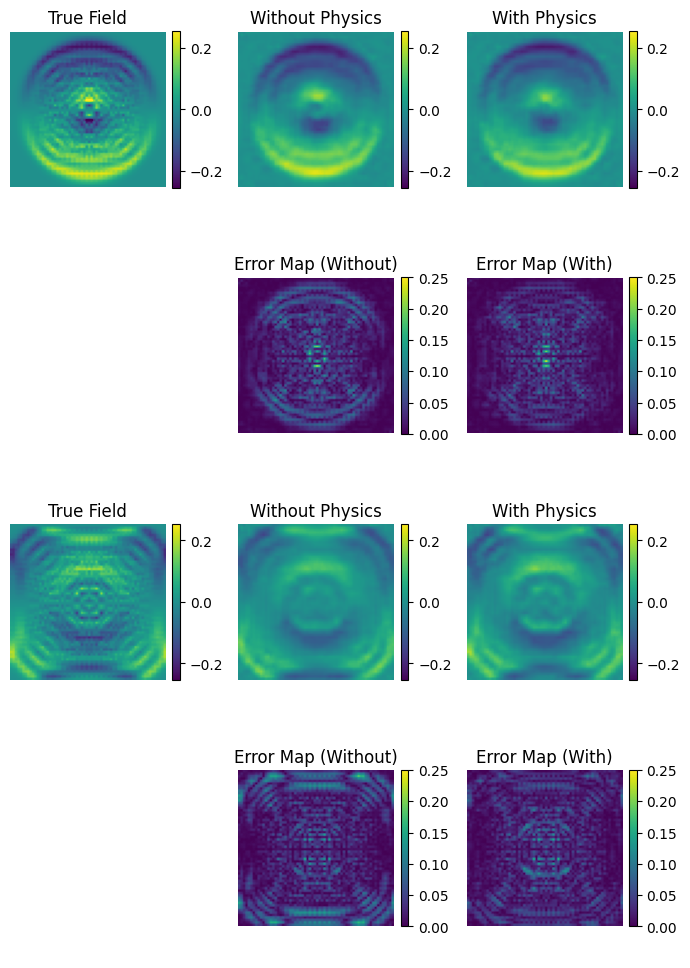}
    \caption{Comparative illustration of the CED-LSTM model's rolling forecast performance at Step 50 and Step 150. The first row shows the true field and predictions with and without physics at Step 50. The second row presents the corresponding error maps. The third row displays the true field and predictions with and without physics at Step 150. The fourth row shows the corresponding error maps.}
    \label{fig:Physical_Comparison}
\end{figure*}

\begin{figure*}[ht!]
    \centering
    \includegraphics[width=\textwidth]{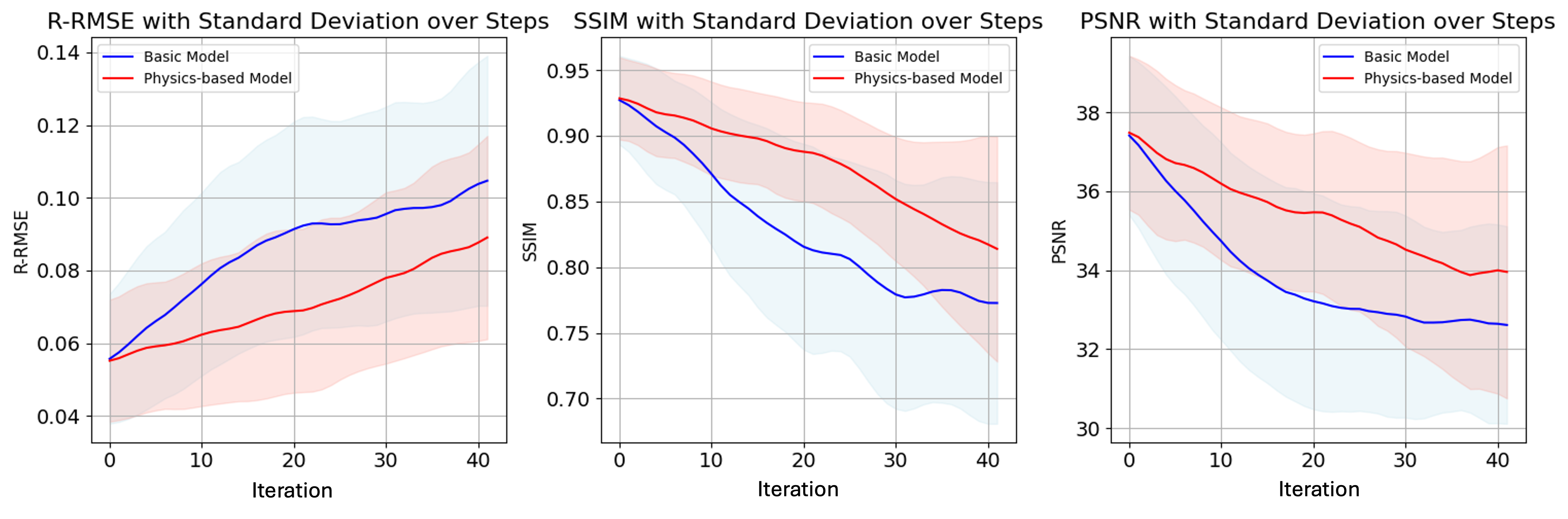}
    \caption{Prediction metric comparison for physics-constrained CED-LSTM model across 42 iterations (210 steps) from Step 75, showing average values (solid lines) and variance (shaded areas) across ten test simulations.}
    \label{fig:prediction_metrics_comparison}
\end{figure*}

Figure~\ref{fig:Physical_Comparison} showcases predictions at Steps 50 and 150 for physics-constrained \ac{CED-LSTM} model. The basic \ac{CED-LSTM} model exhibits a clear decline in its ability to accurately capture evolutionary relationships, with predicted feature representations and local contours increasingly losing definition as the rolling forecasts progress. From the PSNR trends shown in Figure~\ref{fig:prediction_metrics_comparison}, there is a notable decrease in PSNR from approximately 37 dB to 33 dB after 20 iterations, validating the visual degradation observed in Figure~\ref{fig:Physical_Comparison}. In contrast, physics-constrained CED-LSTM, with the implementation of energy conservation constraints, maintains clearer and more stable feature representations even after 30 iterations, as evidenced in the third column of Figure~\ref{fig:Physical_Comparison}.

Further analysis of the \ac{R-RMSE}, \ac{SSIM}, and \ac{PSNR} metrics throughout the rolling forecast sequence consistently shows a performance advantage for physics-constrained CED-LSTM across all evaluation points. As illustrated by the lines and shaded areas in Figure~\ref{fig:prediction_metrics_comparison}, the mean and variance of these metrics (R-RMSE, SSIM, PSNR) across ten test simulations are consistently superior for physics-constrained CED-LSTM compared to the basic CED-LSTM, highlighting its increased stability. Notably, in the 10 test simulations, physics-constrained \ac{CED-LSTM} achieves superior predictive accuracy, with the average SSIM increasing from 0.83 in the basic \ac{CED-LSTM} to 0.88, marking a 5.44\% improvement, and the average PSNR rising from 33.84 dB to 35.47 dB, reflecting a 4.84\% enhancement. The reduction in \ac{R-RMSE} from 0.09 to 0.07 further underscores the enhanced fidelity and quality of rolling forecasts enabled by the integration of physics constraints.
\begin{figure*}[ht!]
    \centering
    \includegraphics[width=\textwidth]{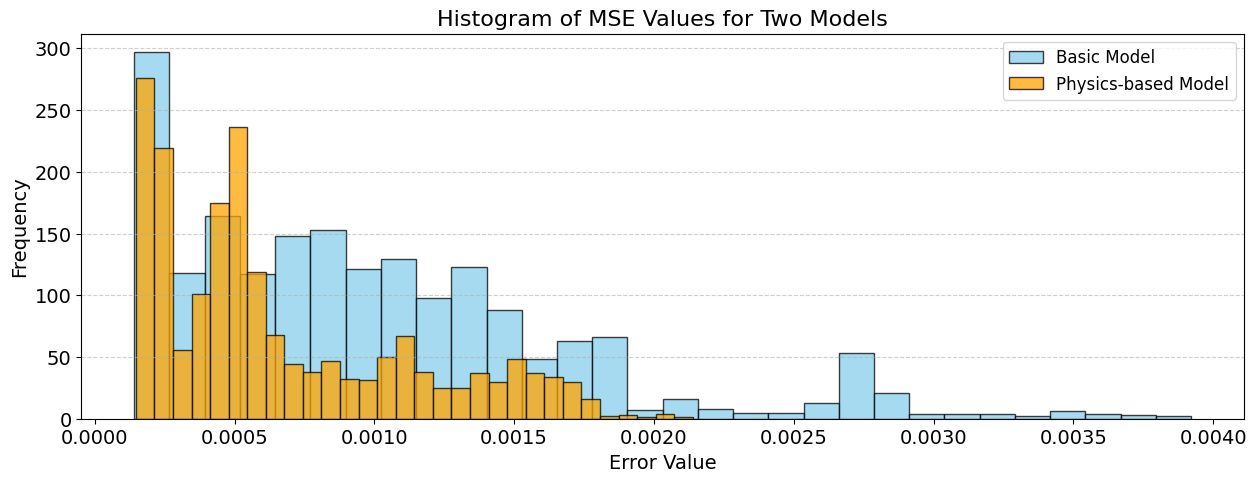}
    \caption{Error histogram comparison in shallow water system with energy conservation constraints using CED-LSTM model.}
    \label{fig:hist_cedlstm_comparison}
\end{figure*}

Additionally, we aggregated the single-step errors (MSE) across the test simulations and constructed a histogram, as shown in Figure~\ref{fig:hist_cedlstm_comparison}. This histogram indicates that the physics-constrained model produces significantly fewer high-error instances compared to the basic model. 

\subsubsection{Effect of \(\lambda_{\text{energy}}\) on Physics-constrained CED-LSTM}

We conducted comparative experiments to evaluate the impact of \(\lambda_{\text{energy}}\) on the CED-LSTM, using values of \(1e-09\), \(2e-09\), and \(1e-10\) for \(\lambda_{\text{energy}}\). The analysis considered the effects of both excessively large and excessively small parameters.

Firstly, for CED-LSTM, as shown in Table~\ref{tab: lambda_energy on CED-LSTM}, a $\lambda_{energy}$ limit in the range of \(1e-10\) to \(2e-09\) can improve the long-term prediction accuracy of the model to some extent. However, as previously mentioned, excessively large weighted parameters interfere with the reduction of the model's primary data-driven loss, as demonstrated in Figures 16(c) and 16(d). This interference ultimately affects the model's overall performance. Conversely, a \(\lambda_{\text{energy}}\) that is too small does not yield optimal results for the physics-based model, as shown in Figures 16(e) and 16(f). Although there is a slight improvement in model accuracy, Figure 16(e) indicates significant variance in predictions, suggesting that an overly small \(\lambda_{\text{energy}}\) might introduce noise during model optimization.

    \begin{table}[h]
    \centering
    \caption{Performance Metrics for Different \(\lambda_{energy}\) Settings on CED-LSTM}
    \begin{tabular}{|c|c|c|c|}
    \hline
    \textbf{\(\lambda_{energy}\)} & \textbf{Metric} & \textbf{Basic} & \textbf{Physics-based} \\ \hline
    \multirow{3}{*}{2e-9} & SSIM & 0.833 & 0.829 \\ \cline{2-4} 
    & PSNR & 33.836 dB & 34.077 dB \\ \cline{2-4} 
    & R-RMSE & 0.086 & 0.082 \\ \hline
    \multirow{3}{*}{1e-9} &   SSIM & 0.832 & 0.859 \\ \cline{2-4} 
    &  PSNR & 33.836 dB & 34.631 dB \\ \cline{2-4} 
    &  R-RMSE & 0.086 & 0.079 \\ \hline
    \multirow{3}{*}{\textbf{5e-10}} &  SSIM & 0.833 & \textbf{0.881} \\ \cline{2-4} 
    &  PSNR & 33.836 dB & \textbf{35.473 dB} \\ \cline{2-4} 
    &  R-RMSE & 0.086 & \textbf{0.069} \\ \hline
    \multirow{3}{*}{1e-10} &  SSIM & 0.833 & 0.844 \\ \cline{2-4} 
    &  PSNR & 33.836 dB & 34.375 dB \\ \cline{2-4} 
    &  R-RMSE & 0.086 & 0.083 \\ \hline
    \end{tabular}
    \label{tab: lambda_energy on CED-LSTM}
    \end{table}

    \begin{figure}[ht]
        \centering
        \begin{subfigure}[b]{0.55\textwidth}
            \includegraphics[width=\textwidth]{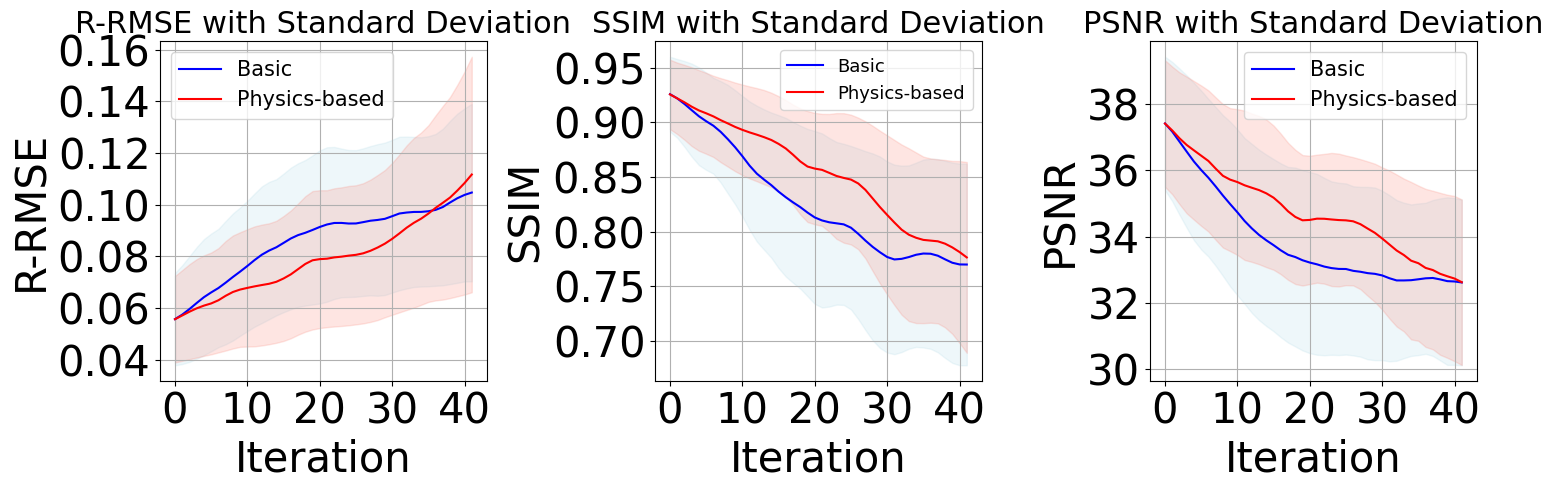}
            \caption{CED-LSTM predictions with \(\lambda_{\text{energy}} = 1e-09\).}
            \label{fig:1e09_ced}
        \end{subfigure}
        \hfill
        \begin{subfigure}[b]{0.4\textwidth}
            \includegraphics[width=\textwidth]{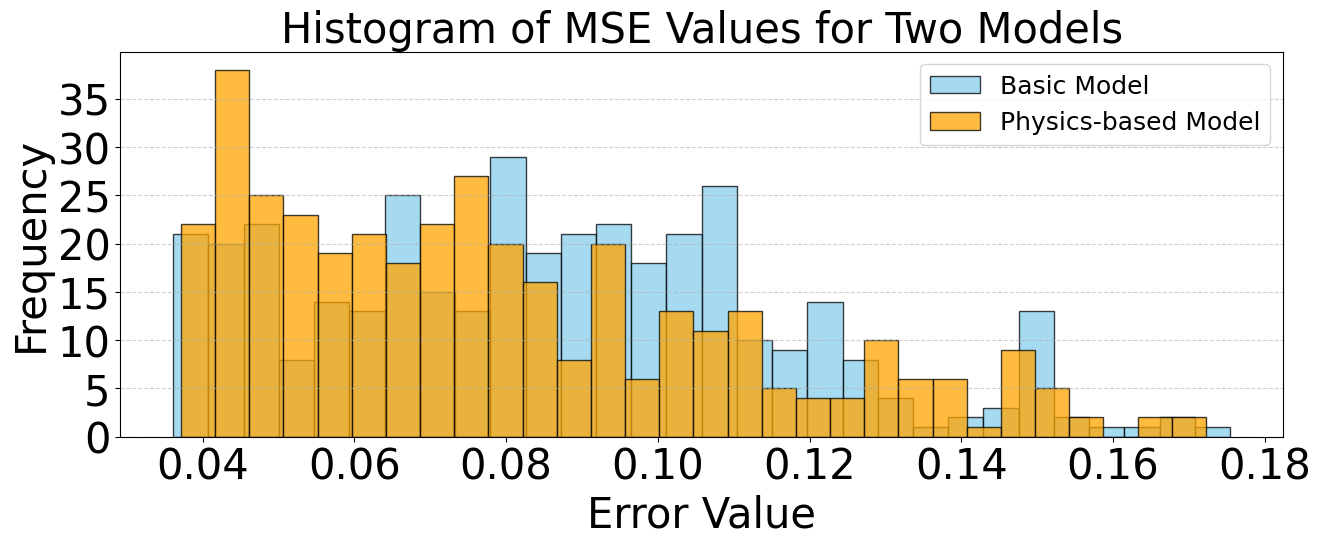}
            \caption{MSE histogram for \(\lambda_{\text{energy}} = 1e-09\).}
            \label{fig:1e09_cedMSE}
        \end{subfigure}

        \begin{subfigure}[b]{0.55\textwidth}
            \includegraphics[width=\textwidth]{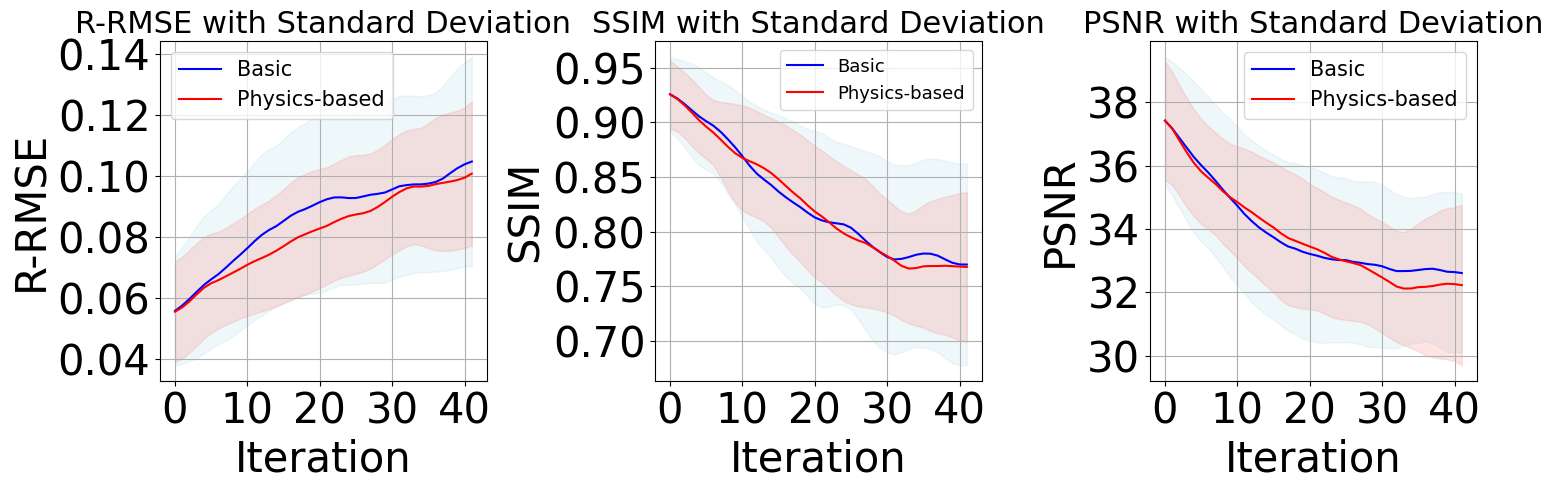}
            \caption{CED-LSTM predictions with \(\lambda_{\text{energy}} = 2e-09\).}
            \label{fig:very_large}
        \end{subfigure}
        \hfill
        \begin{subfigure}[b]{0.4\textwidth}
            \includegraphics[width=\textwidth]{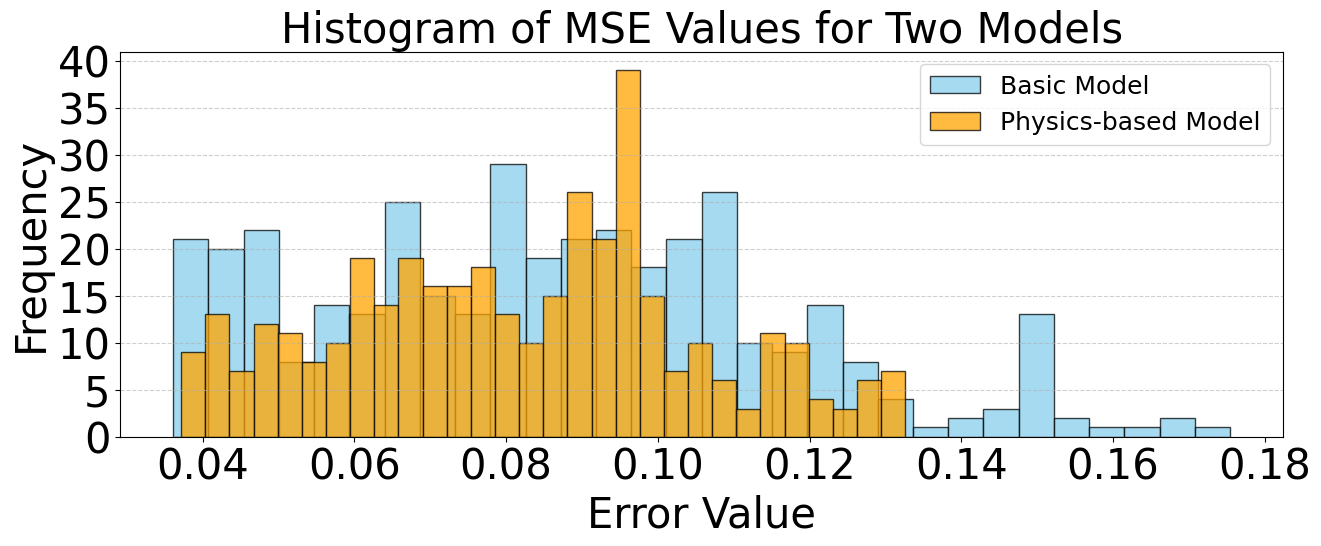}
            \caption{MSE histogram for \(\lambda_{\text{energy}} = 2e-09\).}
            \label{fig:very_large_mse}
        \end{subfigure}

        \begin{subfigure}[b]{0.55\textwidth}
            \includegraphics[width=\textwidth]{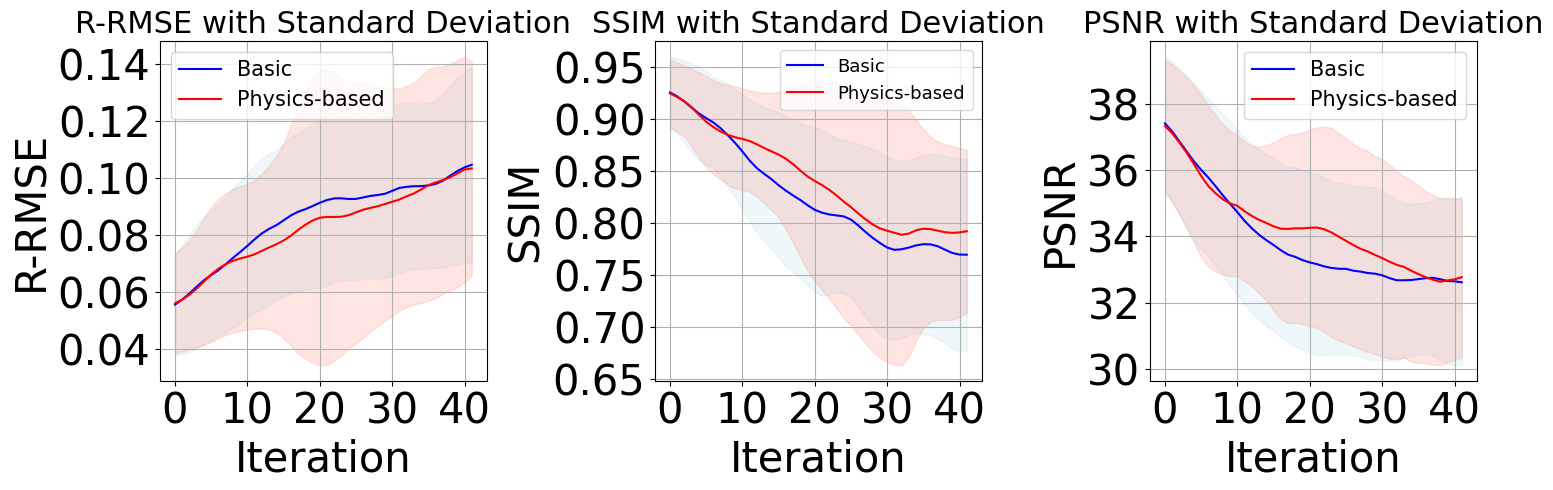}
            \caption{Prediction metric comparison for \(\lambda_{\text{energy}} = 1e-10\).}
            \label{fig:1e10_ced}
        \end{subfigure}
        \hfill
        \begin{subfigure}[b]{0.4\textwidth}
            \includegraphics[width=\textwidth]{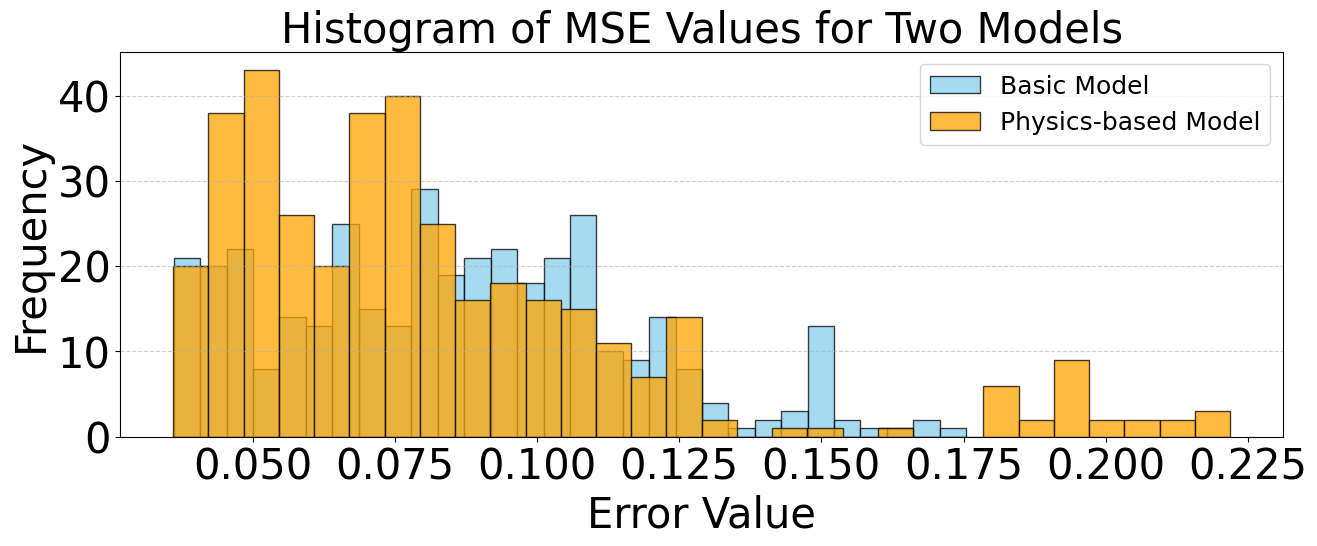}
            \caption{MSE histogram for \(\lambda_{\text{energy}} = 1e-10\).}
            \label{fig:1e10_cedMSE}
        \end{subfigure}

        \caption{Comparison of CED-LSTM predictions and MSE histograms for different \(\lambda_{\text{energy}}\) values.}
        \label{fig:comparison}
    \end{figure}

\subsection{DSOVT (ConvLSTM)}

\subsubsection{Accuracy and Efficiency}
As shown in Table~\ref{tab:SW-model_performance}, despite its computational intensity, ConvLSTM achieves an SSIM of 0.88 and a PSNR of 38.76 dB, marking improvements of 20.55\% in SSIM and 56.23\% in PSNR compared to 2D-Kriging. These advancements underscore ConvLSTM's refined capability in accurately simulating shallow water fields, demonstrating its adeptness in deciphering complex spatial and temporal correlations. Such proficiency ensures that ConvLSTM is particularly suitable for scenarios that require simultaneously simplifying the training process while maintaining prediction accuracy, making it an ideal end-to-end model. However, it is important to note that ConvLSTM has a slightly prolonged inference time of 6.31 seconds due to its large parameter size.

\begin{figure*}[h!]
    \centering
    \includegraphics[width=1.0\textwidth]{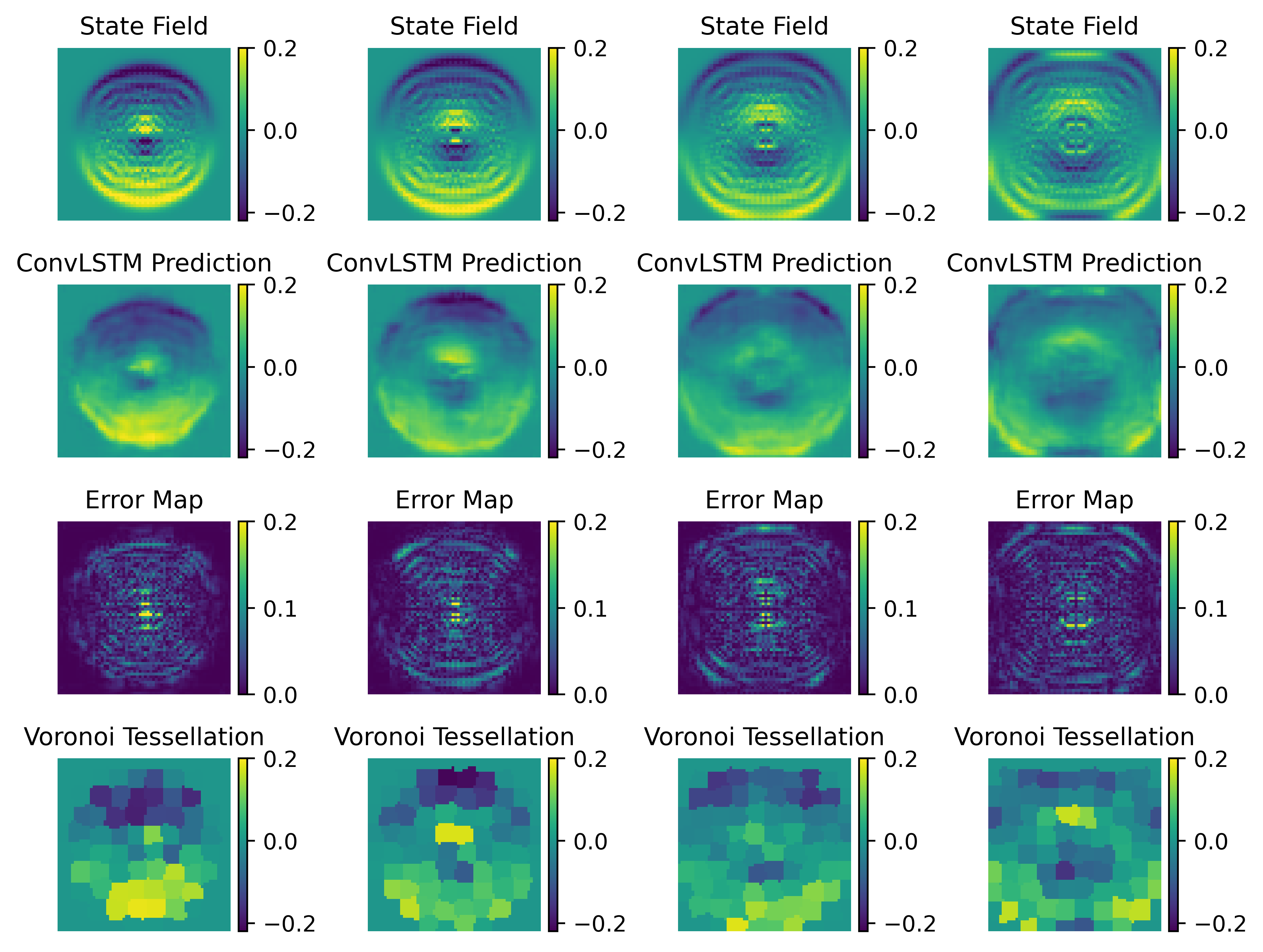}
    \caption{Visualization of \ac{ConvLSTM}'s performance in simulating shallow water dynamics. This figure illustrates the model's predictive journey from input to reality, structured as follows: (1) initial state fields showing baseline conditions of the shallow water system; (2) \ac{ConvLSTM}-predicted fields derived from Voronoi tessellation inputs; (3) error maps underscoring differences between predictions and actual states; and (4) the input Voronoi tessellation for each case. Through four rows and individual columns for unique scenarios, the model's adeptness at capturing fluid dynamics is clearly depicted.}
    \label{fig:ConvLSTM-compare}
\end{figure*}

We can also see from Table~\ref{tab:SW-model_performance} that compared to the 0.96 SSIM and 42.91 dB of CED-LSTM, ConvLSTM does not stand out significantly. From Figure~\ref{fig:ConvLSTM-compare} and Figure~\ref{fig:all-compare}, particularly the Error Maps, it is evident that the largest discrepancies in ConvLSTM's predictions occur at the junctions of water waves within the shallow water field.
As shown in the fourth row of Figure~\ref{fig:ConvLSTM-compare}, the information contained in the Voronoi tessellation inputs is sparse and fails to capture the detailed feature representations of the actual fields. Moreover, with only 30 simulations for training, it is insufficient for the \ac{ConvLSTM} to learn the intricate feature representations present in the true fields. The strength of the \ac{CED} framework is attributed to its preliminary phase of leveraging \ac{CED} spatial predictive capabilities to address some of the missing information from Voronoi tessellation, followed by the use of LSTM for future state predictions. This sequential methodology significantly improves the model's ability to enrich sparse data, thereby enhancing overall predictive accuracy. In contrast, \ac{ConvLSTM} adopts an end-to-end strategy, entering directly Voronoi tessellation to forecast real-world phenomena. 

\begin{table}[ht]
    \centering
    \begin{tabular}{lcccc}
        \hline
        Model & SSIM & PSNR (dB) & R-RMSE & Inference Time (s) \\ \hline
        2D-Kriging & 0.73 & 24.81& 0.27 & 17.43 \\
        3D-Kriging & 0.57 & 20.54& 0.34 & 2961.94 \\
        CED-LSTM & 0.96 & 42.91&0.05 & 0.80 \\
        ConvLSTM & 0.88 & 38.76& 0.07& 6.31 \\
        \hline
    \end{tabular}
    \caption{Comparative assessment of different models' performance in multi-step predictions on testing datasets of shallow water systems, evaluated using SSIM, PSNR (dB), R-RMSE and Inference Time in seconds.}
    \label{tab:SW-model_performance}
\end{table}

\subsubsection{Effect of energy conservation constraints on ConvLSTM}

This study utilizes a small training dataset consisting of the first 10 training simulations, without any alterations to the testing sets, to examine the effects of integrating physics constraints within the ConvLSTM framework. This approach facilitates an evaluation of how physics constraints can boost ConvLSTM's performance in rolling forecasts under conditions of limited data availability. Here, we also choose $\lambda_{\text {energy}} = 5e-10$ as our weight coefficient of the energy constraint for ConvLSTM. In rolling forecasts of ConvLSTM, we also start at step 75 across 10 test simulations. We restrict our discussion to 32 iterations of rolling forecasts for ConvLSTM, as opposed to the 42 iterations used for CED-LSTM. This limitation is due to the significant blurring of inputs received by ConvLSTM after 32 iterations, rendering them nearly devoid of useful information, which makes further comparison potentially meaningless.

\begin{figure*}[h!]
\centering
\includegraphics[width=1.0\textwidth]{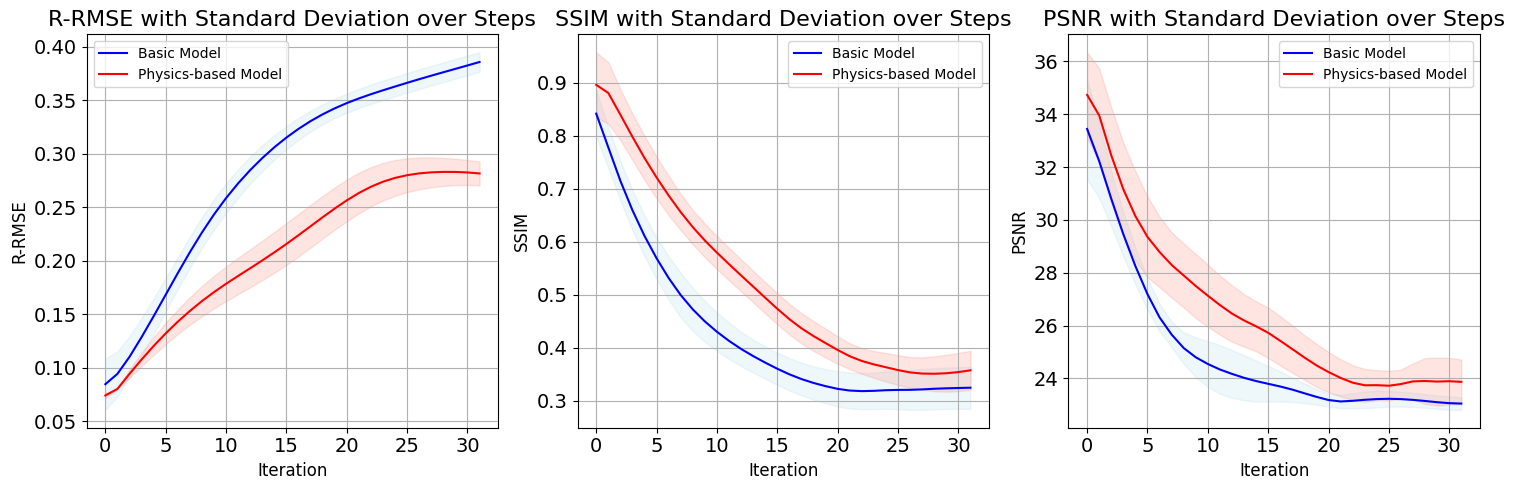}
\caption{Comparison of prediction metrics for the physics-constrained \ac{ConvLSTM} model over 32 iterations (160 steps), starting at the 75th step. Average values are depicted by solid lines across ten test simulations, with shaded areas representing variance among them.}
\label{fig:prediction_metrics_comparison_convlstm}
\end{figure*}

\begin{figure*}[ht!]
    \centering
    \includegraphics[width=\textwidth]{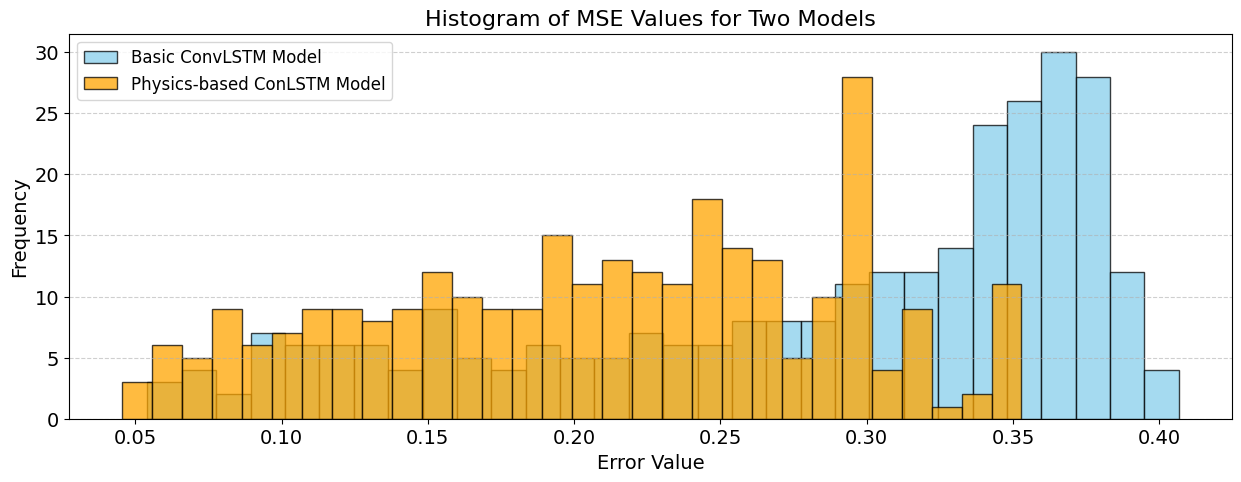}
    \caption{Histogram comparison of error rates in the shallow water system, highlighting the effect of integrating an energy conservation constraints into the \ac{ConvLSTM} model.}
    \label{fig:prediction_hist_cedlstm_comparison}
\end{figure*}

Figures~\ref{fig:prediction_metrics_comparison_convlstm} and~\ref{fig:prediction_hist_cedlstm_comparison} illustrate the progression and error distribution, respectively, for the basic and physics-constrained \ac{ConvLSTM}, focusing on rolling forecasts using Equations~\ref{eq:initial_prediction} and~\ref{eq:rolling_prediction}.
Incorporating physics constraints, specifically energy conservation constraints, results in a substantial reduction in \ac{MSE} and improvements in \ac{SSIM} and \ac{PSNR} metrics compared to the standard \ac{ConvLSTM} approach. Figure~\ref{fig:prediction_metrics_comparison_convlstm} uses shaded areas to illustrate a reduced range of standard deviations, demonstrating how energy conservation constraints diminish prediction errors and enhance both the consistency and reliability of the model.
The implementation of an energy conservation constraint significantly enhances the model's performance, with SSIM increasing by nearly 20.96\% and PSNR by approximately 11.53\%. The R-RMSE also improves, showing a decrease of approximately 26.15\% from 0.28 to 0.21. These improvements underscore the value of integrating physical principles into the training of \ac{ConvLSTM} models.
As shown in the SSIM and PSNR sections of Figure~\ref{fig:prediction_metrics_comparison_convlstm}, physics-constrained \ac{ConvLSTM} exhibits higher accuracy and greater robustness during rolling forecasts compared to the basic model. Similarly, the R-RMSE section in Figure~\ref{fig:prediction_metrics_comparison_convlstm} also demonstrates these characteristics.
Within 10 iterations, the R-RMSE metric increases from 0.08 to about 0.25, the SSIM metric declines sharply from approximately 0.80 to 0.45, and in just 5 iterations, the PSNR decreases from about 34 dB to 27 dB. This trend in \ac{ConvLSTM}'s rolling forecasts can likely be attributed to the inherent design of \ac{ConvLSTM}. Although convolutional layers excel at capturing spatial details, they may struggle to adapt to rapid dynamic changes, especially when the input data comprises sparsely distributed spatial information like Voronoi tessellations. This makes predicting the feature representations of fields over time less effective. Furthermore, \ac{LSTM} tends to propagate errors forward, potentially leading to error accumulation. The interaction between convolutional accuracy and \ac{LSTM} error propagation may be a critical factor in the observed rapid decrease in performance. Nevertheless, the introduction of energy conservation constraints can potentially enhance the model's predictive performance, as indicated by the trends in Figure~\ref{fig:prediction_metrics_comparison_convlstm}. Furthermore, as shown in Figure~\ref{fig:prediction_hist_cedlstm_comparison}, physics-constrained \ac{ConvLSTM} produces fewer large errors in the predictions compared to the basic model. By analyzing the histogram of the MSE, we observe that the frequency of MSEs up to 0.01 is virtually the same for both models. However, the prediction errors of physics-constrained \ac{ConvLSTM} are generally concentrated around 0.25, while those of the basic model are concentrated around 0.37.

\subsubsection{Effect of \(\lambda_{\text{energy}}\) on Physics-constrained ConvLSTM}

As shown in Table~\ref{tab: lambda_energy on ConvLSTM}, the energy constraint improves the long-term prediction accuracy of ConvLSTM within a certain range. Unlike CED-LSTM, ConvLSTM is more sensitive to changes in \(\lambda_{\text{energy}}\). This may be due to the end-to-end pipeline from the sparse noise input of the Voronoi tessellation to the state-field output, which requires the model to learn the physical rules in the image while fitting the image values. Consequently, the model is susceptible to energy constraint loss, achieving energy conservation conditions by generating chaotic outputs.

\begin{figure}[ht]
        \centering
        \begin{subfigure}[b]{0.55\textwidth}
            \includegraphics[width=\textwidth]{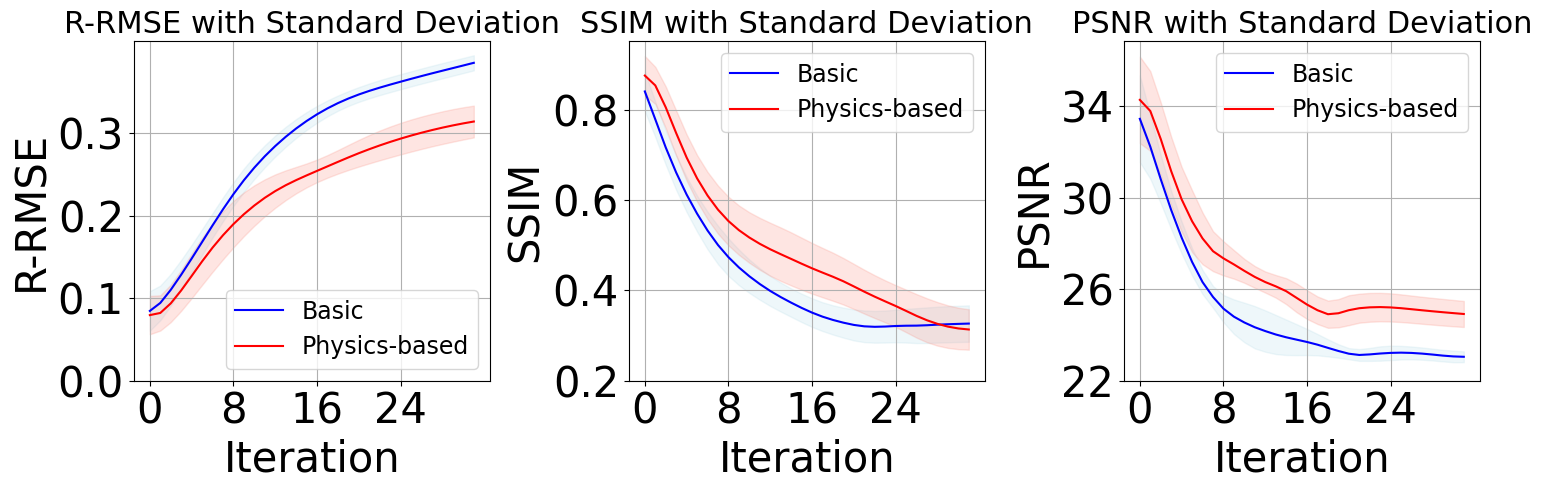}
            \caption{Metrics with \(\lambda_{\text{energy}} = 1e-09\).}
            \label{fig:1e09_convlstm}
        \end{subfigure}
        \hfill
        \begin{subfigure}[b]{0.4\textwidth}
            \includegraphics[width=\textwidth]{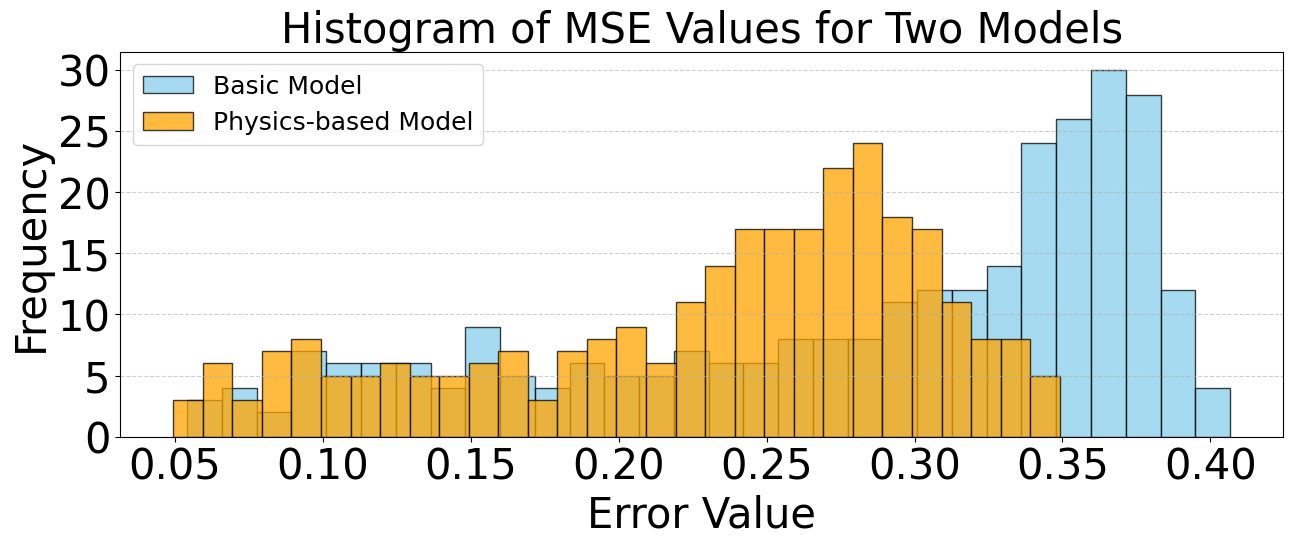}
            \caption{MSE histogram for \(\lambda_{\text{energy}} = 1e-09\).}
            \label{fig:1e09_convlstmMSE}
        \end{subfigure}

        \begin{subfigure}[b]{0.55\textwidth}
            \includegraphics[width=\textwidth]{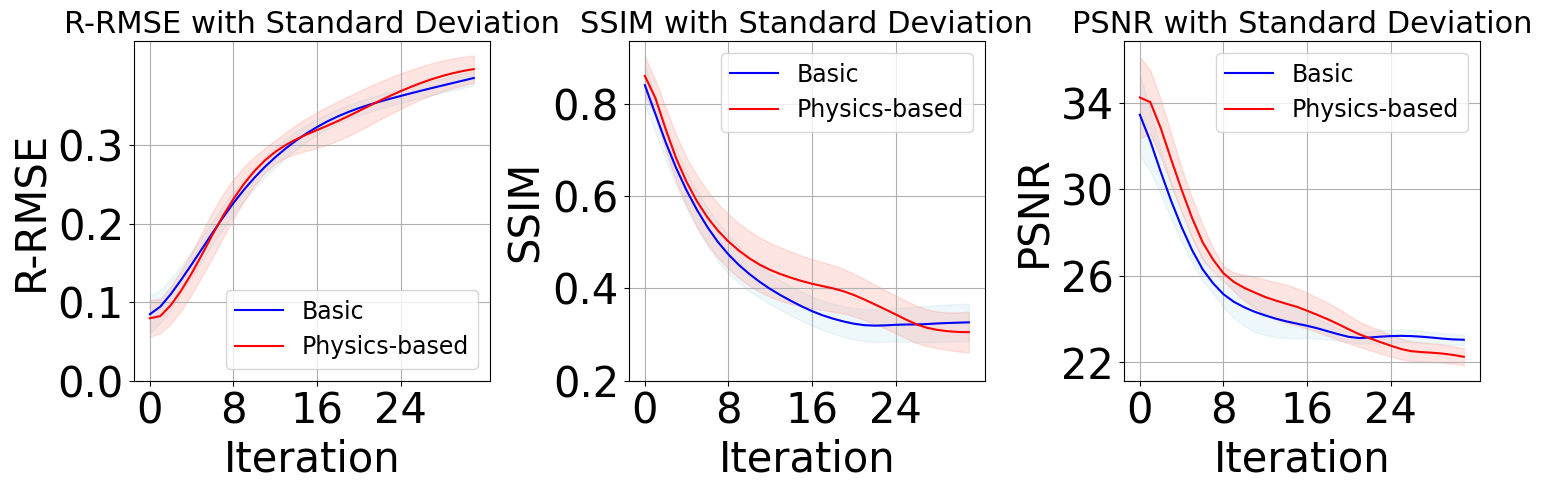}
            \caption{Metrics with \(\lambda_{\text{energy}} = 2e-09\).}
            \label{fig:2e09_ConvLSTM}
        \end{subfigure}
        \hfill
        \begin{subfigure}[b]{0.4\textwidth}
            \includegraphics[width=\textwidth]{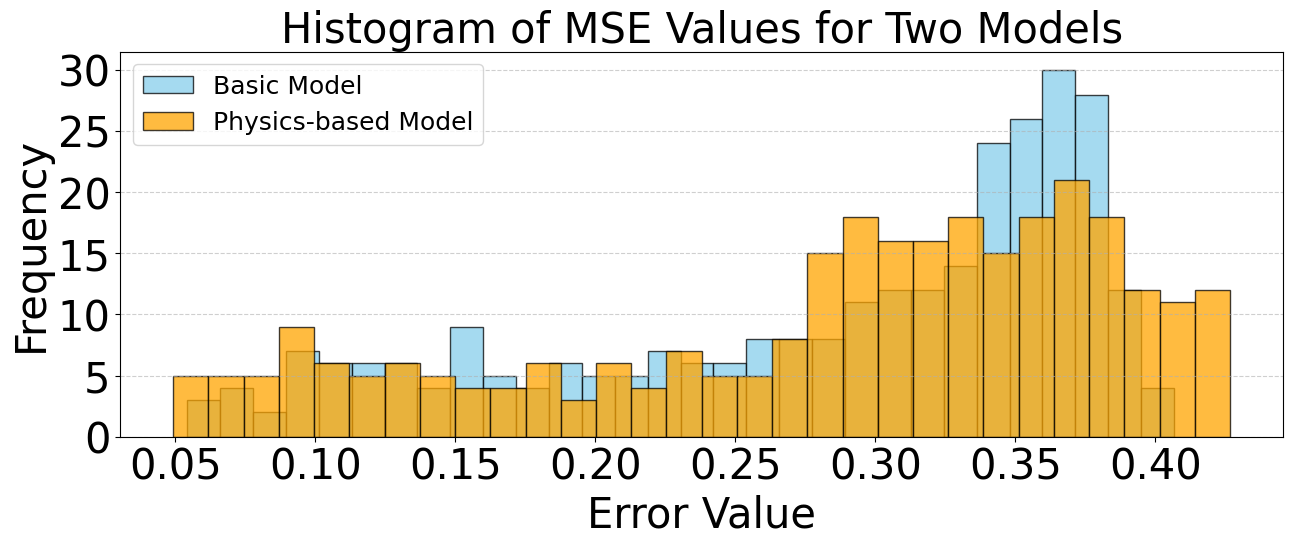}
            \caption{MSE histogram for \(\lambda_{\text{energy}} = 2e-09\).}
            \label{fig:2e09_ConvLSTMmse}
        \end{subfigure}

        \begin{subfigure}[b]{0.55\textwidth}
            \includegraphics[width=\textwidth]{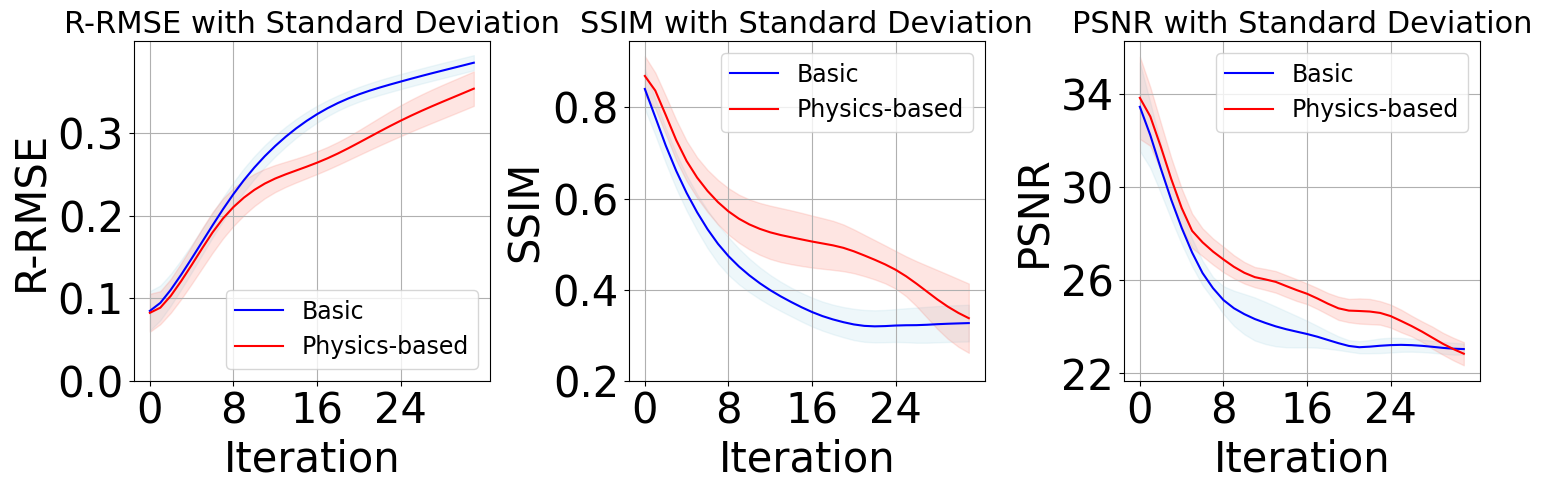}
            \caption{Metrics with \(\lambda_{\text{energy}} = 1e-10\).}
            \label{fig:1e10_ConvLSTM}
        \end{subfigure}
        \hfill
        \begin{subfigure}[b]{0.4\textwidth}
            \includegraphics[width=\textwidth]{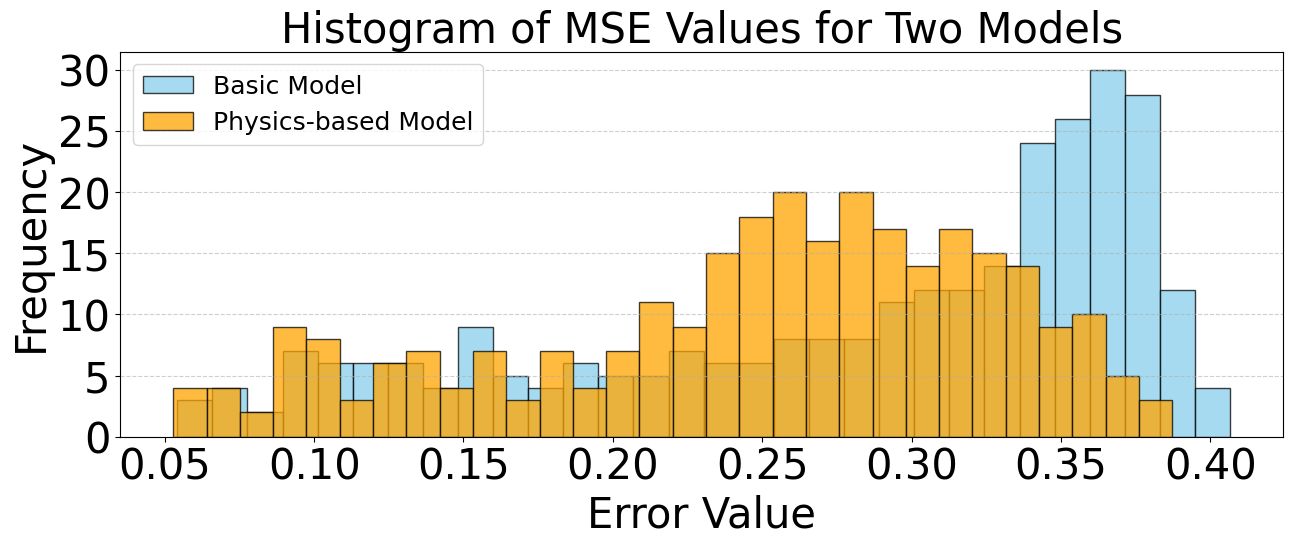}
            \caption{MSE histogram for \(\lambda_{\text{energy}} = 1e-10\).}
            \label{fig:1e10_ConvLSTMMSE}
        \end{subfigure}

        \caption{Comparison of ConvLSTM predictions and MSE histograms for different \(\lambda_{\text{energy}}\) values.}
        \label{fig:convlstm_comparison}
    \end{figure}

Figure~\ref{fig:convlstm_comparison} shows the impact of different \(\lambda_{\text{energy}}\) values on ConvLSTM models. Figures 20(a), 20(c), and 20(e) illustrate the prediction performance of the ConvLSTM model under different \(\lambda_{\text{energy}}\) values. It can be seen that \(\lambda_{\text{energy}} = 1e-09\) and \(\lambda_{\text{energy}} = 2e-09\) show significant improvement in long-term predictions with less variance compared to \(\lambda_{\text{energy}} = 1e-10\). Figures 20(b), 20(d), and 20(f) show the MSE histograms of the ConvLSTM model under different \(\lambda_{\text{energy}}\) values. We can see that the error distributions of Figures 20(b) and 20(f) are almost identical up to an error value of 0.28, but Figure 20(b) has fewer high-error instances compared to Figure 20(f). It is evident that the energy constraint value significantly impacts the model's error distribution, particularly when the \(\lambda_{\text{energy}}\) value is too large or too small, leading to a substantial increase in error variance, exceeding that of the basic model. This demonstrates that ConvLSTM is more sensitive to changes in energy constraints.

    \begin{table}[h]
    \centering
    \caption{Performance Metrics for Different $\lambda_{\text{energy}}$ Settings on ConvLSTM}
    \begin{tabular}{|c|c|c|c|}
    \hline
    \textbf{$\lambda_{\text{energy}}$} & \textbf{Metric} & \textbf{Basic} & \textbf{Physics-based} \\ \hline
    \multirow{3}{*}{2e-9} &  SSIM & 0.415 & 0.471 \\ \cline{2-4} 
    &  PSNR & 23.519 dB & 24.018 dB \\ \cline{2-4} 
    &  R-RMSE & 0.283 & 0.271 \\ \hline
    \multirow{3}{*}{1e-9} &  SSIM & 0.415 & 0.479 \\ \cline{2-4} 
    &  PSNR & 23.519 dB & 25.259 dB \\ \cline{2-4} 
    &  R-RMSE & 0.283 & 0.231 \\ \hline
    \multirow{3}{*}{\textbf{5e-10}} &  SSIM & 0.415 & \textbf{0.502} \\ \cline{2-4} 
    &  PSNR & 23.519 dB & \textbf{26.231 dB} \\ \cline{2-4} 
    &  R-RMSE & 0.283 & \textbf{0.209} \\ \hline
    \multirow{3}{*}{1e-10} &  SSIM & 0.415 & 0.516 \\ \cline{2-4} 
    &  PSNR & 23.519 dB & 24.583 dB \\ \cline{2-4} 
    &  R-RMSE & 0.283 & 0.249 \\ \hline
    \end{tabular}
    \label{tab: lambda_energy on ConvLSTM}
    \end{table}

\section{Conclusion}
\label{sec:Conclusion}

A key innovation of the proposed \ac{DSOVT} framework is its effective handling of unstructured data and prediction of sparse and time-varying fields in multi-step prediction, coupled with the direct integration of physics constraints and data-driven losses into the training processes of \ac{CED-LSTM} and \ac{ConvLSTM} for specific dynamical systems with explicit formulas. This ensures that model optimization transcends purely data-driven methods and aligns with fundamental physical principles, which is crucial for improving the physical interpretability of the latent space and leading to more realistic and accurate predictions of underlying physical principles. Additionally, it offers the flexibility to accommodate a variable number of sensors, enhancing its adaptability to different dynamical systems. In our experiments with real-world \ac{NOAA SST} data and shallow water systems, the framework demonstrates substantial potential to reduce computational resources across more complex and higher-dimensional data sets. Our rolling forecasts in shallow water systems illustrate how physics constraints enhance the stability and accuracy of long-term predictions for the dynamical systems with explicit formulas. The increased accuracy, realism, and stability of predictions are essential for developing effective control strategies for dynamical systems, ensuring that the predictions remain consistent with physical laws and directly impacting the efficiency of control actions.

Given its adaptability and computational efficiency, \ac{DSOVT} is exceptionally well-suited for real-time forecasting tasks in environmental and industrial fluid dynamics. However, the framework has its limitations, particularly in \ac{ConvLSTM}'s predicting feature representations in the state fields. 
The challenge with \ac{ConvLSTM} arises when dealing with sparse or incomplete data, particularly when the input lacks sufficient data size, spatio-temporal continuity, and detail. To address this, we are exploring the incorporation of depthwise separable convolutions into \ac{ConvLSTM} to reduce the model’s parameters and enhance its width and depth, thereby boosting its ability to predict sparse data and feature representations~\cite{pfeuffer2019separable}. Furthermore, we plan to integrate self-attention mechanisms, especially leveraging the Transformer model architecture, to enhance the model’s ability to capture long-term dependencies~\cite{li2020bidirectional}. Correspondingly for CED-LSTM, inspired by the works of Fukami et al.~\cite{fukami2024data} on extreme aerodynamics and the data-driven approach to transient lift attenuation and Fukami and Taira~\cite{fukami2023grasping} on refining the fluid dynamics model to better capture physics, improving interpretability and accuracy for advanced air vehicles in tough weather, we recognize the importance of retaining physical meaning within the latent vectors. We plan to incorporate techniques that ensure the latent space maintains a direct correspondence to the underlying physics of the fluid dynamics, allowing for more interpretable and reliable predictions.


\section*{Data and code availability}
The code of the shallow water experiments is available at \url{https://github.com/EdWangLoDaSc/DSOVT}. Sample data and the script to generate experiments are also provided in the repository.


\section*{Acronyms}

\begin{acronym}[AAAAA]
\footnotesize{
\acro{NN}{Neural Network}
\acro{VAR}{Vector Autoregressive}
\acro{MSE}{Mean Squared Error}
\acro{R-RMSE}{Relative Root Mean Squared Error}
\acro{VCNN}{Voronoi-based Convolutional Neural Network}
\acro{RBF}{Radial Basis Function}
\acro{ML}{Machine Learning}
\acro{CNN}{Convolutional Neural Networks}
\acro{SSIM}{Structural Similarity Index Measure}
\acro{PSNR}{Peak Signal-to-Noise Ratio}
\acro{AE}{Autoencoder} 
\acro{CAE}{Convolutional Autoencoder} 
\acro{CED}{Convolutional Encoder-Decoder}
\acro{CED-LSTM}{Convolutional Encoder-Decoder combined with Long Short-Term Memory}
\acro{ConvLSTM}{Convolutional Long Short-Term Memory}
\acro{LSTM}{Long Short-Term Memory}
\acro{POD}{Proper Orthogonal Decomposition}
\acro{DSOVT}{Dynamical System Prediction from Sparse Observations using Voronoi Tessellation}
\acro{ROM}{Reduced-Order Modelling}
\acro{DL}{Deep Learning}
\acro{GNNs}{Graph Neural Networks}
\acro{NOAA SST}{National Oceanic and Atmospheric Administration Sea Surface Temperature}
\acro{PDEs}{Partial Differential Equations}
\acro{2D-Kriging}{Two-Dimensional Linear Kriging Regressions}
\acro{3D-Kriging}{Three-Dimensional Spatio-Temporal Kriging Interpolation}
\acro{ReLU}{Rectified Linear Unit}
\acro{tanh}{hyperbolic tangent}
\acro{SINDy}{Sparse Identification of Nonlinear Dynamical Systems}
}
\end{acronym}

\footnotesize{
\bibliographystyle{elsarticle-num-names}
\bibliography{main.bib}}
\end{document}